\documentclass[letterpaper,anonymous]{article} 
\usepackage{aaai23}  
\usepackage{times}  
\usepackage{helvet}  
\usepackage{courier}  
\usepackage[hyphens]{url}  
\usepackage{graphicx} 
\usepackage{natbib}  
\usepackage{caption} 
\usepackage{algorithm}
\usepackage{algorithmic}
\usepackage{url}
\usepackage{comment}
\usepackage{natbib}
\usepackage{caption}
\usepackage{subcaption}
\usepackage{multirow}
\usepackage[np, autolanguage]{numprint}
\usepackage{enumitem}
\usepackage[utf8]{inputenc}
\usepackage{soul}
\usepackage{graphicx}
\usepackage{booktabs}
\usepackage[flushmargin]{footmisc}
\usepackage{xr}

\makeatletter
\newcommand*{\addFileDependency}[1]{
\typeout{(#1)}
\@addtofilelist{#1}
\IfFileExists{#1}{}{\typeout{No file #1.}}
}\makeatother

\newcommand*{\myexternaldocument}[1]{%
\externaldocument{#1}%
\addFileDependency{#1.tex}%
\addFileDependency{#1.aux}%
}
\myexternaldocument{supp_material_hcomp23}

\usepackage{newfloat}
\usepackage{listings}
\DeclareCaptionStyle{ruled}{labelfont=normalfont,labelsep=colon,strut=off} 
\floatstyle{ruled}
\newfloat{listing}{tb}{lst}{}
\floatname{listing}{Listing}
\title{Collect, Measure, Repeat: \\Reliability Factors for Responsible AI Data Collection}

\author {
    Oana Inel,\textsuperscript{\rm 1}
    Tim Draws,\textsuperscript{\rm 2}
    Lora Aroyo\textsuperscript{\rm 3}
}
\affiliations {
    \textsuperscript{\rm 1} University of Zurich, Zurich, Switzerland\\
    \textsuperscript{\rm 2} Delft University of Technology, Delft, the Netherlands\\
    \textsuperscript{\rm 3} Google Research, NY, NY, USA\\
    inel@ifi.uzh.ch, t.a.draws@tudelft.nl, l.m.aroyo@gmail.com
}

\begin{document}

\maketitle
\begin{abstract}
The rapid entry of machine learning approaches in our daily activities and high-stakes domains demands transparency and scrutiny of their fairness and reliability. To help gauge machine learning models' robustness, research typically focuses on the massive datasets used for their deployment, \emph{e.g.}, creating and maintaining documentation for understanding their origin, process of development, and ethical considerations. However, data collection for AI is still typically a one-off practice, and oftentimes datasets collected for a certain purpose or application are reused for a different problem. Additionally, dataset annotations may not be representative over time, contain ambiguous or erroneous annotations, or be unable to generalize across issues or domains. Recent research has shown these practices might lead to unfair, biased, or inaccurate outcomes. We argue that data collection for AI should be performed in a responsible manner where the quality of the data is thoroughly scrutinized and measured through a systematic set of appropriate metrics. In this paper, we propose a Responsible AI (RAI) methodology designed to guide the data collection with a set of metrics for an iterative in-depth analysis of the \emph{factors influencing the quality and reliability} of the generated data. We propose a granular set of measurements to inform on the \emph{internal reliability} of a dataset and its \emph{external stability} over time. We validate our approach across nine existing datasets and annotation tasks and four content modalities. This approach impacts the \emph{assessment of data robustness} used for AI applied in the real world, where diversity of users and content is eminent. Furthermore, it deals with fairness and accountability aspects in data collection by providing systematic and transparent quality analysis for data collections.

\end{abstract}



\maketitle

\section{Introduction}
\label{sec:introduction}

As the use of machine learning (ML) and artificial intelligence (AI) becomes more ubiquitous in our daily activities, \emph{e.g.}, to pick a restaurant for dinner~\cite{burke2002hybrid}, as well as in high-stakes domains, \emph{e.g.}, to select a job candidate~\cite{li2021algorithmic} or choose medical treatment for a patient~\cite{shatte2019machine}, the need to scrutinize every aspect of AI systems is also increasing. This includes evaluating their training and testing data quality, as well as quantifying the level of fairness, transparency, accountability, and non-maleficence \cite{jobin2019global} these systems have. Several actionable toolkits and checklists for both models and datasets have been proposed, such as Fairlearn \cite{bird2020fairlearn}, AI Fairness 360 \cite{bellamy2019ai}, Aequitas \cite{saleiro2018aequitas}, Model Cards \cite{mitchell2019model}, Datasheets for Datasets \cite{gebru2021datasheets}, PAIR AI Explorables,\footnote{https://pair.withgoogle.com/explorables/} AI Test Kitchen.\footnote{https://blog.google/technology/ai/join-us-in-the-ai-test-kitchen/} Furthermore, this also led to an emerging data-centric research effort on how data quality can affect the robustness, reliability, and fairness of AI systems' performance in the real world~\cite{mehrabi2021survey,sambasivan2021everyone,kapania2020data}. 

Traditionally, high-quality data for ML is collected from experts and inter-rater reliability (IRR) scores (\textit{e.g.}, Cohen's $\kappa$~\cite{cohen1960coefficient}, Fleiss' $\kappa$~\cite{fleiss1971measuring}, or Krippendorff's $\alpha$~\cite{krippendorff2011computing}) measure their reliability. Employing experts, however, is often costly and time-consuming. Crowdsourcing is a widely used alternative to create ground truth datasets for ML applications. Due to the nature of crowdsourcing annotation studies (\emph{i.e.}, raters who likely have limited or no domain expertise), a large body of research has primarily focused on data evaluation and aggregation techniques~\cite{hovy2013learning,dumitrache2018crowdtruth,paun2018comparing,braylan2020modeling}. 

Under the assumption that each annotated input sample has only one correct interpretation~\cite{nowak2010reliable}, crowdsourced annotations are typically aggregated using majority vote (MV) \cite{dumitrache2021empirical}. However, research has shown that data quality is complex and can be influenced by many \emph{factors}, such as disagreement-prone or subjective tasks, ambiguous input samples, target annotations, and guidelines, diverse rater characteristics and perspectives~\cite{welinder2010online,aroyo2014threesides,kairam2016parting,Chang:2017:Revolt,draws2022effects}, ethical aspects and power structures in annotation processes~\cite{miceli2020between,diaz2022crowdworksheets}, or cognitive biases \cite{eickhoff2018cognitive,santhanam2020studying,draws2022effects}. In such cases, IRR scores may not always be able to capture the true annotations' reliability and MV could eliminate correct answers vetted by only a few raters. Additionally, IRR scores cannot be used to directly compare datasets, as they only indicate raters' consistency rather than data quality. 

These factors have led to several streams of research. First, the notion of ground truth currently adopts a perspectivist stance~\cite{basile2021toward,aroyo2015truth} which highlights the need of diverse opinions and perspectives for a better knowledge representation compared to MV. Second, before runtime checklists have been proposed~\cite{draws2021checklist,thomas2022crowd} to help researchers consider cognitive biases and other human factors affecting their annotations. Third, attention is drawn to formulating dataset artifacts that describe the collection purpose, method, and raters~\cite{bender2018data,ramirez2020drec,gebru2021datasheets,diaz2022crowdworksheets}. Fourth, existing datasets have been extensively judged, improved, and re-annotated based on empirical evidence suggesting that existing annotations are not representative anymore or contain ambiguous or erroneous annotations~\cite{yun2021re,inel2019validation,aroyo2015truth}. 

\begin{figure}[ht!]
\centering
    \includegraphics[width=0.475\textwidth]{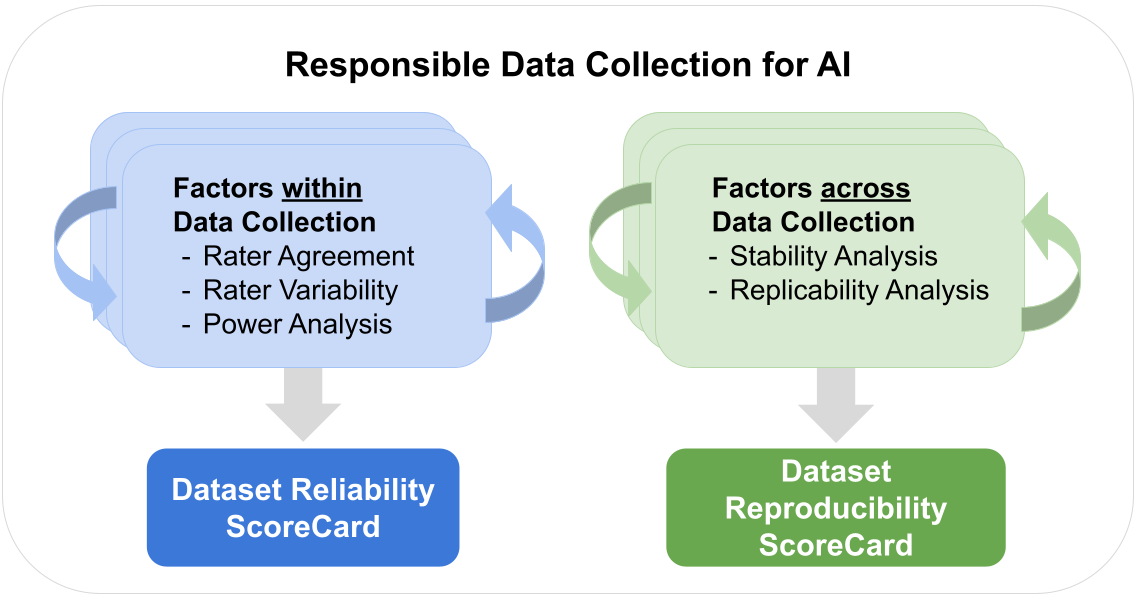}
    \caption{Methodology for measuring reliability and reproducibility of AI data collections.}
    \label{fig:framework}
\end{figure}

However, the research landscape is still lacking a unified framework that allows for cross-datasets comparisons and measurement of dataset stability for repeated data collections. Thus, or proposed approach complements existing research by proposing an \emph{iterative metrics-based methodology} for thoroughly scrutinizing the \emph{factors} influencing the intrinsic reliability of datasets and their stability over time and for various contexts or factors (Fig. \ref{fig:framework}). 

Our proposed methodology enables a comprehensive analysis of data collections by applying reliability and reproducibility measurements in a systematic manner. The reliability metrics are applied at the level of a single data collection and focus on understanding the raters. We then propose that data collection campaigns are repeated, either under similar or different conditions. This allows us to study in-depth the reproducibility of the datasets and their stability under various conditions or constraints, using a set of reproducibility metrics. In short, we propose a set of metrics that are applied (1) on a single repetition and (2) across repetitions to thoroughly evaluate the reliability and stability of annotated datasets. The overall methodology is designed to integrate responsible AI practices into data collection for AI. This allows data practitioners to follow our step-by-step guide to explore factors influencing reliability and quality, ensuring transparent and responsible data collection practices. We validate our methodology on nine existing data collections repeated at different time intervals with similar or different rater qualifications. The annotation tasks span different degrees of subjectivity, data modalities (text and videos), and data sources (Twitter, search results, product reviews, YouTube videos). 

The following are the key contributions of this paper:\footnote{Supplemental material and analysis at: \url{https://github.com/oana-inel/ResponsibleAIDataCollection}}

\begin{enumerate}[noitemsep,nosep,topsep=0pt,leftmargin=*]
    \item a step-wise guide for practitioners consisting of a set of metrics to thoroughly investigate and explore factors that influence or impact the reliability of human-annotated datasets;
    \item a validation and illustration of the proposed metrics-based iterative methodology for achieving transparency of the reliability of datasets and their stability over time on nine existing data collections; and
    \item a discussion of implications and lessons learned for responsible data collection practices.
\end{enumerate}

\section{Related Work on Data Excellence}
\label{sec:related_work}

Data quality is an already established field of study \cite{wang1996beyond,zaveri2016quality,lukyanenko2015information}, spanning numerous domains and use cases, such as linked open data, user-generated data, to name a few. Data quality aspects are also addressed in several ISO standards (e.g., ISO 25012~\cite{international2008iso,guerra2023iso}, ISO 8000\footnote{https://www.iso.org/standard/81745.html}). Research utilizing crowdsourced datasets has further broadened the community's views regarding factors influencing or affecting data quality. A large body of research has been focusing on acknowledging the impact of cognitive biases, such as confirmation bias or anchoring effect, on the crowdsourced data quality~\cite{eickhoff2018cognitive,hube2019understanding,santhanam2020studying,draws2022effects}. However, while various types of cognitive biases are known to impact data quality, in typical annotation studies, it is not a mainstream practice to account for raters' stances, opinions, or knowledge on various issues. Furthermore, despite several proposed mechanisms to mitigate biases~\cite{eickhoff2018cognitive,hube2019understanding,barbosa2019rehumanized}, it is still unclear which mechanisms are suitable for a particular situation or which individual characteristics of the raters may lead to systematic biases~\cite{draws2022effects}. To further alleviate some of these issues,~\citet{draws2021checklist} proposed a 12-item checklist for requesters to identify which cognitive biases might affect their data before the start of data collection. While this checklist offers a powerful tool for requesters, many times AI and ML practitioners reuse existing annotated datasets that might lack proper documentation or description of the annotation process, which makes the assessment difficult and leads to worrying outcomes when deployed in the real world \cite{paullada2021data}.

Unequal distribution of demographic characteristics among raters may subsequently lead to poor performance of ML models~\cite{barbosa2019rehumanized}. While investigating whether different cultural communities produce different gold standards and whether algorithms perform differently on gold standards from different cultural communities, \citet{sen2015turkers} found that AMT-derived gold standards for knowledge-oriented tasks can not generalize across different communities and influence ML performance. In the context of image annotation, \citet{dong2012war} concluded that different cultures provide different tags, being highly influenced by their cognitive and emotional aspects. The same behavior was observed when performing news sentiment analysis~\cite{DBLP:conf/lrec/BalahurSKZGHPB10}. Besides cultural differences, cognitive biases, and stereotypes, societal events or temporal aspects can also add variance in collected data~\cite{aroyo2015truth,christoforou2021s,sen2015turkers}. \citet{christoforou2021s} showed empirical evidence that significant public health events might be reflected in the descriptive tags regarding a person's identity and body weight that raters use to annotate images of people. 

In the remainder of this section, we review three streams of research regarding responsible data collection for AI: collection, assessment, and documentation and maintenance.

\subsection{Data Collection}
\label{subsec:responsible_data_collection}

Data collection practices strongly affect the quality of crowdsourced data. This led to extensive explorations in evaluating raters' performance and identifying under-performing pools of raters~\cite{ipeirotis2010quality,bozzon2013reactive,soberon2013measuring}, improving the overall clarity of the annotation task design~\cite{Kittur:2008:CUS:1357054.1357127,gadiraju2017clarity,wu2017confusing,han2019all}, encouraging raters to reflect on their answers~\cite{kutlu2020annotator}, or experimenting with several annotation designs to identify the most suitable one to capture the appropriate answers~\cite{inel2018study,lau2014measuring,roitero2018fine}. More precisely, at the level of the task design, many studies experimented with annotation scales. For example,~\citet{roitero2018fine} showed that fine-grained annotation scales are more suitable and natural than coarse-grained annotation scales to capture web documents' relevance. Intrinsic motivation and incentives have also been shown to be beneficial in improving the quality of crowdsourced data~\cite{ho2015incentivizing,kittur2013future}. In our research, we present extensive data analysis of nine existing data annotation tasks which vary in terms of collection criteria such as input data (tweets, product reviews, web documents, facial expression recordings, and news broadcasts) and annotation goal.

\subsection{Data Assessment}
\label{subsec:data_assessment}

To measure the reliability of crowdsourced annotations, research focused on quality control mechanisms \cite{daniel2018quality} and the definition of aggregation techniques \cite{hung2013evaluation,li2019exploiting,dumitrache2018crowdtruth,hovy2013learning,paun2018comparing}.
Typically, current annotation campaigns rely on the use of multiple raters per annotated input and reporting of inter-rater reliability metrics \cite{park2012crowdsourcing,sigurdsson2016much,park2014toward}, such as Cohen's $\kappa$~\cite{cohen1960coefficient}, Fleiss' $\kappa$~\cite{fleiss1971measuring}, or Krippendorff's $\alpha$~\cite{krippendorff2011computing}. However, the choice of the IRR metric is less important than having a representative number of raters per input~\cite{artstein2008inter}. Furthermore, \citet{popovic2022reporting} found that raw counts are a more suitable input for computing and estimating inter-rater reliability compared to normalized counts or percentages in a machine translation use case. In a recent study, however, \citet{braylan2022measuring} point out that Krippendorff's $\alpha$ relies on mean distances, which can lead to mistakenly discarding good data when dealing with subjective annotations and propose using more suitable distance functions depending on the task at hand. When dealing with subjective tasks, or tasks that could generate diverse opinions or perspectives and potentially multiple ground truths, achieving reliable results is thus even more challenging \cite{graham2018evaluation,7042476,basile2021toward}. This could mean that different replications of such a task could give very different results. 

To the best of our knowledge, only a few data collection experiments addressed repeatability \cite{blanco2011repeatable,DBLP:journals/corr/abs-1911-01875}. Thus, our work proposes data collection repeatability as a responsible practice to measure data stability over time. We propose a set of metrics carefully chosen to scrutinize the human factors influencing various aspects of the data over time, thus fostering cross-comparison between datasets.

\subsection{Data Documentation and Maintenance}
\label{subsec:data_documentation}

By systematically reviewing 150 published papers dealing with classification tasks on Twitter data, \citet{geiger2020garbage} concluded that issues such as reliability, transparency, and accountability in data collection practices are not a mainstream approach in the ML community. A considerable amount of analyzed papers offer limited or no details regarding raters, their demographic information, compensation details, IRR scores of the collected datasets, annotation instructions, and overall setup of the annotation process. Following studies that also showed the extent to which potential biases are present in extensively used image datasets~\cite{DBLP:conf/emnlp/ZhaoWYOC17,DBLP:conf/eccv/HendricksBSDR18,otterbacher2015crowdsourcing}, a lot of attention has been brought to data documentation and maintenance approaches, in many fields. 

Inspired by medicine and psychology literature, \citet{bender2018data} proposed data statements for characterizing and understanding the raters of a natural language dataset, their potential biases, and how they might affect the deployment of ML models. Similarly, informed by the electronics industry where every component is thoroughly described in terms of characteristics, test results, or recommended usage, \citet{DBLP:journals/corr/abs-1803-09010} proposed \textit{datasheets} for datasets. \citet{diaz2022crowdworksheets} studied ethical considerations that affect the annotation of the dataset, such as, for instance raters' previous experience, and developed the CrowdWorkSheet framework to facilitate critical reflection and transparent documentation of dataset annotation decisions, processes, and outcomes. \citet{pushkarna2022data} proposed \emph{data cards} to record key aspects of datasets and their life cycle (\emph{i.e.}, explanations concerning the provenance, representation, usage, and fairness of ML datasets for all stakeholders), allowing for responsible AI development. Finally,~\cite{wilkinson2016fair} proposed a set of guiding principles to support researchers, industry practitioners, and funding and publishing agencies in scholarly data reuse; these guiding, measurable principles tackle four fundamental data principles --- findability, accessibility, interoperability, and reusability (i.e., FAIR). In the crowdsourcing community, \citet{ramirez2020drec,ramirez2021state} provide guidelines for requesters to improve dataset reporting and reproducibility. 

While our work is not focusing on documenting the annotation process of human-annotated datasets, it complements existing approaches by proposing a set of reliability analysis metrics that foster responsible data documentation and adherence to proper data provenance and documentation guidelines for subsequent dataset alterations.

\section{Reliability and Reproducibility Metrics for Responsible Data Collection}
\label{sec:reliability_analysis}

In this section, we introduce our proposed methodology for in-depth analysis of the \emph{reliability} and \emph{reproducibility} of data annotation studies. The proposed methodology brings together, in a systematic way, a set of measurements typically performed in an ad-hoc manner. However, the ability to observe their interaction allows data practitioners to provide a holistic picture of the data quality produced by these studies. Therefore, our proposed methodology provides a step-wise approach as a guide for practitioners to explore factors that influence or impact the reliability and quality of their collected data. While the reliability analysis focuses on the raters that participate in a data collection, the reproducibility analysis provides insights regarding the stability of the overall dataset. As observed in Figure~\ref{fig:framework}, the chosen metrics provide input for a scorecard allowing for thorough and systematic evaluation and comparison of different data collection experiments. 

Ultimately, the proposed reliability and reproducibility scorecards and analyses allow for more transparent and responsible data collection practices. This leads to the identification of factors that influence quality and reliability, the thorough measurement of dataset stability over time or in different conditions, and allows for datasets comparison.

\subsection{Measuring the reliability of data annotations}

We address the reliability of the crowdsourced annotations by looking at the \emph{raters agreement}, \emph{rater variability}, and \emph{power analysis} to determine the sufficient number of raters needed for each task. These analyses equip us with fundamental observations and findings for characterizing the quality and reliability of the annotations. It is important to note that depending on the nature and characteristics of the task (\emph{i.e.}, difficulty, subjectivity, clarity), the assessment of the crowdsourced annotations' reliability is not always trivial and needs to be considered when generalizing the results across different tasks. 

\paragraph{Rater agreement analysis:} indicates the level of consistency among raters' annotations in an experiment. We compute the inter-rater reliability score (IRR) using Krippendorff's $\alpha$~\cite{krippendorff2011computing} because is suitable for most annotation experiments, given that it can deal with multiple raters, various rating types (\emph{i.e.}, categorical, ordinal, interval), and missing data (\emph{i.e.}, not all units are annotated by all raters). Typically, $\alpha$ scores above 0.8 are considered to show strong or high agreement among raters, while values close to 0.6 are still considered acceptable~\cite{landis1977measurement,carletta1996assessing,krippendorff1980validity}. Note that the IRR scores can be influenced by various characteristics of the task, which need to be taken into consideration in the overall analysis. For example, a low IRR reliability score (e.g., below or very close to 0.33) could indicate both high task subjectivity and unsuitable annotation guidelines or raters' qualifications, among others. Interpreting whether the agreement value is low, medium, or high, however, is often task-dependent and should be discussed on a case basis. 

\paragraph{Rater variability analysis:} gives insights regarding the variability in raters' answers distributions. We use two metrics to measure raters' precision for each individual annotated item in our datasets - we inspect the standard deviations in raters' annotations~\cite{DBLP:journals/corr/abs-1911-01875} when having binary or continuous value annotations and the index of qualitative evaluation (IQV)~\cite{wilcox1967indices} when having nominal or categorical values. IQV is a measure for assessing the variability of nominal variables with values between 0 (all raters’ answers are in one category) and 1 (raters’ answers are evenly distributed in each category). In our analysis, we consider IQV $\leq$ .33 as low variability, .33 $<$ IQV $>$ .66 as medium, and IQV $\geq$ .66 as high.

\paragraph{Power analysis:} indicates whether the number of raters used in each annotation task is sufficient. To identify the optimal number of raters (\emph{i.e.}, for which the variability in terms of IRR is not significant), we bootstrap the number of raters [3,4,5,..,n] per input item, where n is the maximum number of raters~\cite{snow2008cheap}. For each number of raters, we perform 100 runs, where raters are randomly selected for each input item, and each time we compute the IRR. Then, we apply a chi-squared test for one standard deviation and test whether the standard deviation of the IRR scores for each number of raters is lower or equal to a threshold. In our experiments, we considered the threshold of $.01$ and for each number of raters, we test the following hypotheses: $H_{0}: \sigma \leq .01$ and $H_{a}: \sigma > .01$, where $\sigma$ refers to the standard deviation of the IRR scores. Given $H_{0}$, we conduct a right-tailed test, and we search for the lowest number of raters for which we fail to reject $H_{0}$, \emph{i.e.}, p $<$ .05.

\subsection{Measuring annotation reproducibility} To investigate how rater populations influence the reliability of the annotation results, we propose \emph{repeating} the annotations at different time intervals and in different settings, thus identifying the factors influencing their reliability. However, just by using the aforementioned reliability measures, we can not perform a proper comparison of the collected annotations. For instance, high IRR values in several repetitions indicate highly homogeneous raters' populations within each repetition, but it does not necessarily mean that the experiments are highly reproducible. For this, we perform two additional measurements: 1) stability analysis and 2) replicability analysis to understand how much variability the raters bring and how much we can generalize the results. For example, a high correlation between two annotation task repetitions indicates that our results are stable, and the populations that participated in the two repetitions are drawn from the same distribution.

\paragraph{Stability analysis:} is the degree of association of the aggregated raters' scores across two annotation task repetitions. We measure the stability of the data collection with the correlation of the aggregated raters' annotations between pairwise repetitions of each task. The aggregated raters' annotation can be a mean value, majority vote, or any aggregation technique suitable for the task at hand. To compute the correlation, we use the Spearman's rank correlation \cite{xiao2016using} for tasks with numerical values and Chi-square test of independence for tasks with categorical values. 

\paragraph{Replicability similarity analysis:} indicates the degree of agreement between two rater pools, making two data annotation tasks comparable. It consists of raters' agreement across repetitions of a particular task. To measure this, we use a metric called cross-replication reliability (xRR)~\cite{DBLP:conf/acl/WongPA20}. The xRR score between two repetitions is goes from 0 to the highest IRR score of the two repetitions. In a perfect replication, the xRR score is equal to the IRR score of each individual repetition (implying that the two repetitions also have equal IRR scores). This further means that similar IRR and xRR scores indicate both internal and external validity while much lower xRR values compared to IRR scores indicate low external validity.

\section{Published Annotation Tasks and Datasets}
\label{sec:experimental_methodology}
We outline the experimental design to evaluate the reliability and reproducibility of nine published data annotation studies covering a wide range of content modalities and annotation tasks (Table \ref{tab:overview_crowd_experiments}). We first outline the data annotation studies (Section~\ref{sec:crowdsourcing_tasks}), and then describe the resulting datasets (Section~\ref{sec:dataset}). Thus, we use these nine data annotation studies as a two-fold objective: 1) to illustrate how practitioners should apply the methodology introduced in Section~\ref{sec:experimental_methodology} to gain a better understanding of their data and 2) to validate the usefulness of the methodology to help practitioners explore factors that influence or impact data reliability. 

\begin{table*}[!ht]
\caption{Overview of annotation tasks and their settings in terms of input data and annotation template.}
\label{tab:overview_crowd_experiments}
\resizebox{1\textwidth}{!}{
\begin{tabular}{cccccccccccccc} \toprule
\multirow{2}{*}{\begin{tabular}[c]{@{}c@{}}Task \\ Type\end{tabular}} & \multirow{2}{*}{\begin{tabular}[c]{@{}c@{}}Dataset \\ Name\end{tabular}} & \multicolumn{3}{c}{Input} & & \multicolumn{2}{c}{Template} & & \multicolumn{5}{c}{Repetitions} \\ \cmidrule{3-5} \cmidrule{7-8} \cmidrule{10-14}
 & & Modality & Type & \#HITs & &\begin{tabular}[c]{@{}c@{}}Annotation \\ Guidelines\end{tabular} & \begin{tabular}[c]{@{}c@{}}Annotation \\ Value\end{tabular} & & \begin{tabular}[c]{@{}c@{}}\# \\Total\end{tabular} & \begin{tabular}[c]{@{}c@{}}Temporal \\ Distance\end{tabular} & \begin{tabular}[c]{@{}c@{}}Rater \\ Pool\end{tabular} & \begin{tabular}[c]{@{}c@{}}Rater \\ Platform\end{tabular} & \begin{tabular}[c]{@{}c@{}}Repeating \\ Raters?\end{tabular} \\ \midrule
\multirow{5}{*}{\begin{tabular}[c]{@{}c@{}}Video Concepts \\Relevance\end{tabular}} & VCR\_ALL                                                           & \multirow{5}{*}{video} & \multirow{5}{*}{\begin{tabular}[c]{@{}c@{}}news \\ broadcasts\end{tabular}} & 88  &   & \multirow{5}{*}{\begin{tabular}[c]{@{}c@{}}Select all \\ from list\end{tabular}} & Relevant concepts                                                            &    & \multirow{5}{*}{3} & \multirow{5}{*}{> 3 months}                               & \multirow{5}{*}{same} & \multirow{5}{*}{same} & \multirow{5}{*}{yes} \\
& VCR\_E & & & 19   &  &   & Relevant events & & & & \\
 & VCR\_P & & & 23  &   & & Relevant people & & & & \\
 & VCR\_L & & & 22  &   & & Relevant locations & & & & \\
 & VCR\_O & & & 9  &    & & Relevant organizations & & & & \\ \midrule
\begin{tabular}[c]{@{}c@{}}Video Human \\ Facial Emotions\end{tabular}                    & IRep                                                              & video                  & \begin{tabular}[c]{@{}c@{}}human facial \\ recordings\end{tabular}          & 1090  & & \begin{tabular}[c]{@{}c@{}}Select all \\ from list\end{tabular}                  & Facial expressions                                                              & & 4                  & -                                                            & \begin{tabular}[c]{@{}c@{}}same \end{tabular}  & \begin{tabular}[c]{@{}c@{}}same \end{tabular}  & no                                                            \\ \midrule
\begin{tabular}[c]{@{}c@{}}Product Reviews\end{tabular}                              & PR                                                                & text                   & \begin{tabular}[c]{@{}c@{}}product \\ review\end{tabular}                   & 20  &   & \begin{tabular}[c]{@{}c@{}}Select one option \\ from list\end{tabular}           & Product issue                                                                  &  & 5                  & 1 week                                                       & same     & same                                                        & no                                                            \\ \midrule
\begin{tabular}[c]{@{}c@{}}Crisis Tweets\end{tabular}                       & CT                                                                 & text                   & \begin{tabular}[c]{@{}c@{}}Twitter \\ messages\end{tabular}                 & 20   &  & \begin{tabular}[c]{@{}c@{}}Select one option \\ from list\end{tabular}           & Crisis category                                                                  & & 5                  & 1 week                                                       & same  & same                                                           & no                                                            \\ \midrule
\begin{tabular}[c]{@{}c@{}}Words Similarity\end{tabular}                          & WS353                                                            & text                   & word pairs                                                                  & 353  &  & \begin{tabular}[c]{@{}c@{}}Rate the similarity \\ of two words\end{tabular}      & \begin{tabular}[c]{@{}c@{}}Value from 0 to 10,\\ increments of 0.25\end{tabular} & & 3                  & 20 years                                                     & different             & different                                           & no      \\ \bottomrule                                                     
\end{tabular}
}
\end{table*}

\subsection{Annotation Tasks}
\label{sec:crowdsourcing_tasks}
We first describe how the datasets (see Section~\ref{sec:dataset}) used in our experiments have been collected, \emph{i.e.}, the task, the number of repetitions, and their settings (summarized in Table~\ref{tab:overview_crowd_experiments}). All tasks and datasets have already been published.

\textbf{Video Concepts Relevance (VCR)} ~\cite{inel2022fine}. The raters were asked to watch a video of 1-2 minutes and then select all relevant concepts for the content of the video from a list of machine-extracted concepts (an average of 11 concepts). Five different annotation experiments were run, each focusing on the identifications of different concept types, e.g \emph{event} (\textbf{VCR\_E}), \emph{people} (\textbf{VCR\_P}), \emph{location} (\textbf{VCR\_L}), and \emph{organization} (\textbf{VCR\_O}), and concepts of any type (\textbf{VCR\_ALL}). The task was run on Amazon Mechanical Turk (AMT) with ten videos representing short English news broadcasts from YouTube annotated by 15 raters. Each task was repeated three times, at least three months apart, and each repetition used the same raters' qualifications, and raters were allowed to participate across repetitions.

\textbf{Video Human Facial Expressions (IRep)} \cite{DBLP:conf/acl/WongPA20}. The raters were given a video recording containing human facial expressions and were asked to select all facial expression labels (\emph{i.e.}, emotions) that they perceived as being relevant from a predefined list of facial expression labels~\cite{DBLP:conf/acl/WongPA20}. The task was run on AMT. A total of 30 emotion labels (from the set defined by~\cite{cowen2017self}) were shown, together with the option ``unsure'' (raters were instructed to choose this option when it was not possible to determine the facial expressions expressed in the video recording). Each video recording was annotated by two raters. The task was repeated three times, each time with raters from a different pool, namely raters from Mexico City, Kuala Lumpur, Budapest, and internationals. 

\textbf{Product Reviews (PR)} \cite{qarout2019platform}. The raters were given a product review and were asked to classify the issue described in the review into one of three possible classes (\emph{i.e.}, ``size aspects'', ``fit aspects'', ``no issue with size or fit''). The task was run on AMT, and each rater was required to annotate all 20 product reviews, which appeared in the same order for each rater, and each product review was annotated by at least 68 raters. The task was repeated five times at intervals of one week. The raters were not allowed to participate in more than one repetition.

\textbf{Crisis Tweets (CT)} \cite{qarout2019platform}. The raters were given a crisis-related Twitter message and were asked to categorize it into one of nine possible options (\emph{i.e.}, ``injured or dead people'', ``other useful information'', ``infrastructure and utilities damage'', ``not related or irrelevant'', ``sympathy and emotional support'', ``donation needs or offers or volunteering services'', ``missing, trapper or found people'', ``displaced people and evacuations'', ``caution and advice''). The task was run on AMT, and each rater was required to annotate all 20 tweets which appeared in the same order for all raters, and each tweet was annotated by at least 68 raters. The task was repeated five times at intervals of one week. Each rater was allowed to participate in just one repetition.

\textbf{Words Similarity (WS353)} \cite{finkelstein2001placing,DBLP:journals/corr/abs-1911-01875}. The raters were given a pair of words and were asked to rate the similarity of the two words on a scale from 0 to 10 (0 indicating the words are totally unrelated and 10 indicating the words are very closely related) \cite{finkelstein2001placing,DBLP:journals/corr/abs-1911-01875} (fractional scores such as .25, .5, and .75 are also possible). The task was first run by \citet{finkelstein2001placing}, and each pair of words was annotated either by 13 or 16 raters, and each rater annotated all pairs. The second time the task was run by \cite{DBLP:journals/corr/abs-1911-01875} in 2019 (thus around 20 years apart), on AMT. In this repetition, each pair of words was annotated by 13 raters, and each rater was allowed to annotate as many pairs as they wanted.

\subsection{Datasets}
\label{sec:dataset}

The tasks described in Section \ref{sec:crowdsourcing_tasks} resulted in nine annotated datasets covering different data modalities (text and videos of various lengths and duration) and sources (Twitter, product reviews, YouTube), as described in Table~\ref{tab:datasets} and below.

\begin{table}[!ht]
\caption{Overview of datasets used in our experiments.}
\label{tab:datasets}
\centering
\resizebox{0.47\textwidth}{!}{
\begin{tabular}{lllll} \toprule
Dataset                       & \begin{tabular}[c]{@{}c@{}}Input \\Modality\end{tabular} & Content                 & Size & Task Type                           \\ \midrule
VCR\_E & video    & video - event pairs & 208   & Video Concept Rel. \\
VCR\_P & video    & video - people pairs & 234   & Video Concept Rel. \\
VCR\_L & video    & video - location pairs & 223   & Video Concept Rel. \\
VCR\_O & video    & video - organization pairs & 59   & Video Concept Rel. \\
VCR\_ALL & video    & video - concept pairs & 969  & Video Concept Rel. \\
IRep           & video    & human facial recordings & 1065 & \begin{tabular}[c]{@{}c@{}}Video Human Facial \\Expressions\end{tabular}  \\
PR          & text     & product reviews         & 20   & Product Reviews     \\
CT            & text     & Twitter crisis messages & 20   & Crisis Tweets    \\
WS353               & text     & WordNet word pairs      & 353  & Words Similarity    \\ \bottomrule
\end{tabular}
}
\end{table}

\textbf{Video Concept Relevance (VCR\_E, VCR\_P, VCR\_L, VCR\_O, VCR\_ALL)}: Dataset of 208, 234, 223, 59, and respectively 969 video - concept pairs which have been annotated in terms of relevance in the data annotation tasks VCR\_E, VCR\_P, VCR\_L, VCR\_O, and, respectively VCR\_ALL. The concepts were machine-extracted (video subtitles and video stream) from ten short English news broadcasts (\emph{i.e.}, videos) published on YouTube, from a publicly available dataset~\cite{DBLP:conf/um/InelTA20,jong2018human,inel2022fine}. 

\textbf{Video Human Facial Expressions (IRep)}: Dataset of 1090 video recordings of human facial recordings, part of the International Replication (IRep) dataset\footnote{\url{https://github.com/google-research-datasets/replication-dataset}}, published by~\citet{DBLP:conf/acl/WongPA20}. Each video recording is annotated with emotions from 30 available emotions. The video recordings were generally very short, 5 seconds on average (a more extensive description of the recordings is found in~\cite{cowen2017self}).

\textbf{Product Reviews (PR)}: Dataset of 20 English product reviews for fashion items (accompanied by a photo representative of the respective product), randomly selected from the dataset published by \citet{DBLP:journals/corr/abs-1805-09648}. Each product review is annotated with one of three possible issue classes, as described in Section \ref{sec:crowdsourcing_tasks}.

\textbf{Crisis Tweets (CT)}: Dataset of 20 English crisis-related Twitter messages (\emph{e.g.}, earthquake, flood), randomly selected from the dataset published by \citet{DBLP:conf/lrec/ImranMC16}. Each tweet is annotated with one of nine possible crisis-related options, as described in Section \ref{sec:crowdsourcing_tasks}.

\textbf{WordSim (WS353)}\footnote{\url{https://aclweb.org/aclwiki/WordSimilarity-353_Test_Collection_(State_of_the_art)}}: Dataset of 353 English word pairs~\cite{finkelstein2001placing}, used as benchmark for semantic similarity~\cite{witten2008effective} and word embeddings~\cite{levy2014neural,bojanowski2017enriching,pennington2014glove}. The word pairs were selected from WordNet, and include the 30 noun pairs from~\cite{miller1991contextual}. Each pair is annotated in terms of how similar the two words are on a 1 to 10 scale.

\section{Results}
\label{sec:results}

In this section, we report on the results of the reliability analysis of the data collection studies described in Section~\ref{sec:crowdsourcing_tasks}, and in Table~\ref{tab:overview_crowd_experiments}, and their repetitions. We apply our iterative metrics-based evaluation methodology to the nine datasets from these studies. In the analysis of the results, we denote each repetition as $R_{x}$, where $x$ is the repetition index. 

\subsection{VCR: Annotation Tasks and Datasets}
\label{subsec:results_vcr}

We first report on the reliability analysis of the VCR datasets (\emph{i.e.}, VCR\_ALL, VCR\_E, VCR\_P, VCR\_L, VCR\_O), as depicted in the first five rows in Table \ref{tab:agreement_analysis}, columns $R_{1}$, $R_{2}$, and $R_{3}$. We observe that the datasets of all VCR annotation tasks and repetitions have mostly fair agreement and less often moderate agreement ($R_{1}$ \& $R_{3}$ for VCR\_ALL, $R_{1}$ for VCR\_E, and $R_{1}$ \& $R_{2}$ for VCR\_P). The tasks VCR\_O and VCR\_L have, overall, the lowest inter-rater reliability. Similarly, the precision of the annotations in all tasks and repetitions is not substantial. In Table \ref{tab:reliability_individual_replication} in the Appendix, we show an overview of the variability of each repetition of the \emph{VCR} annotation tasks. For each video-concept pair, we computed the standard deviation of their score. The majority of the experiments have a mean standard deviation (MSTD) of around 0.3, with the task \emph{VCR\_O} having a higher value of around 0.36. The standard deviation of deviations (STDD) is similar across tasks and repetitions, with the lowest value observed for the \emph{VCR\_O} task. These high values observed for MSTD and STDD show that this task for annotating relevant concepts in videos is subjective and raters consistently disagree. Concepts of type \emph{organization} seem to generate the most disagreement among raters. 

In our power analysis, we observe that all repetitions of each task tend to display similar variability in terms of IRR score. For each repetition of every task, according to the right-tailed Chi-squared test, we got very similar results in terms of the optimal number of raters needed to annotate a video-concept pair. According to the Chi-square test, the following number of raters is optimal (with minimal variation across repetitions): \emph{VCR\_ALL} - 6 raters, \emph{VCR\_E} - 12 raters, \emph{VCR\_P} - 11 raters, \emph{VCR\_L} - 11 raters, \emph{VCR\_O} - 14 raters. These, in addition to the high variability of IRR scores shown in Figure \ref{fig:bootstrap_workers} in the Appendix, suggest that annotating the relevance of \emph{organizations} in videos is a more difficult task that might require an even larger number of raters.

Although the IRR scores of the annotations gathered in all repetitions are rather low, the Spearman's rank correlation between the relevance score of the video-concept pairs (computed as the ratio of raters that picked the concept as relevant) in each pair of repetitions is high, above 0.85 for all tasks and repetitions, showing a statistically significant, strong positive correlation in Table \ref{tab:reliability_pairwise}. Furthermore, the pairwise xRR scores (see Figures~\ref{fig:vcr_all_xrr}, \ref{fig:vcr_e_xrr}, \ref{fig:vcr_p_xrr}, \ref{fig:vcr_l_xrr}, \ref{fig:vcr_o_xrr}) are very similar to the IRR scores of the repetitions. Thus, we observe that while the IRR scores are rather low, raters are similarly consistent in each repetition and across repetitions, showing that disagreement seems to be intrinsic to the task.

\begin{table}[!ht]
\caption{Rater agreement for all datasets, where the rater agreement is computed with Krippendorff's $\alpha$.}
\label{tab:agreement_analysis}
\centering
\resizebox{0.3\textwidth}{!}{
\begin{tabular}{lccccc}
\toprule
         & $R_{1}$   & $R_{2}$   & $R_{3}$   & $R_{4}$   & $R_{5}$   \\ \midrule
VCR\_ALL & 0.43 & 0.40 & 0.44 & -    & -    \\
VCR\_E   & 0.44 & 0.37 & 0.41 & -    & -    \\
VCR\_P   & 0.41 & 0.40 & 0.39 & -    & -    \\
VCR\_L   & 0.30 & 0.38 & 0.34 & -    & -    \\
VCR\_O   & 0.25 & 0.30 & 0.30 & -    & -    \\
IRep     & 0.25 & 0.23 & 0.50 & 0.13 & -    \\
PR       & 0.41 & 0.33 & 0.32 & 0.20 & 0.36 \\
CT       & 0.59 & 0.70 & 0.68 & 0.72 & 0.65 \\
WS353    & 0.59 & 0.57 & 0.50 & -    & -   \\ \bottomrule
\end{tabular}
}
\end{table}

\begin{table*}[!ht]
\centering
\caption{Spearman's $\rho$ rank correlation of the relevance of each video-concept pair for each pair of VCR tasks repetitions.}
\resizebox{0.8\textwidth}{!}{
\label{tab:reliability_pairwise} 
\begin{tabular}{cccccc} \toprule
 & VCR\_ALL & VCR\_E & VCR\_P & VCR\_L & VCR\_O \\ \cmidrule{2-6}

R1 \& R2 & \begin{tabular}[c]{@{}c@{}}$\rho$=0.90, $p$=0.0 \end{tabular} & \begin{tabular}[c]{@{}c@{}}$\rho$=0.90, $p$=5.05e-76 \end{tabular} & \begin{tabular}[c]{@{}c@{}}$\rho$=0.91, $p$=4.72e-89 \end{tabular} & \begin{tabular}[c]{@{}c@{}}$\rho$=0.91, $p$=9.24e-86 \end{tabular} & \begin{tabular}[c]{@{}c@{}}$\rho$=0.87, $p$=3.86e-19 \end{tabular}  \\

R1 \& R3 & \begin{tabular}[c]{@{}c@{}}$\rho$=0.90, $p$=0.0 \end{tabular} & \begin{tabular}[c]{@{}c@{}}$\rho$=0.89, $p$=1.97e-72 \end{tabular} & \begin{tabular}[c]{@{}c@{}}$\rho$=0.90, $p$=4.25e-85 \end{tabular} & \begin{tabular}[c]{@{}c@{}}$\rho$=0.89, $p$=4.06e-77 \end{tabular}  & \begin{tabular}[c]{@{}c@{}}$\rho$=0.86, $p$=1.05e-18 \end{tabular} \\

R2 \& R3 & \begin{tabular}[c]{@{}c@{}}$\rho$=0.90, $p$=0.0 \end{tabular} & \begin{tabular}[c]{@{}c@{}}$\rho$=0.90, $p$=2.60e-75 \end{tabular} & \begin{tabular}[c]{@{}c@{}}$\rho$=0.91, $p$=5.25e-85 \end{tabular} & \begin{tabular}[c]{@{}c@{}}$\rho$=0.91, $p$=4.50e-86 \end{tabular}  & \begin{tabular}[c]{@{}c@{}}$\rho$=0.85, $p$=8.77e-18 \end{tabular}  \\ \bottomrule 
\end{tabular}
}
\end{table*}

\subsection{IRep: Annotation Task and Dataset}
\label{subsec:results_irep}

\citet{DBLP:conf/acl/WongPA20} already provide an in-depth analysis of the IRR and xRR scores per emotion in three of the repetitions of the tasks. More precisely, they analyze the agreement among raters from three different regions, i.e., Mexico City ($R_1$), Kuala Lumpur ($R_2$), and Budapest ($R_3$). Their main conclusion is that raters seem to have more similar or divergent agreement values depending on their country of origin. In terms of individual emotions, they also observe that the most or least agreed-upon emotions are different for each country. Similarly, only a few emotions seem to have both internal and external validity, as the xRR scores between pairwise repetitions indicate.  

In addition, in this paper, we also analyze $R_4$, a repetition of the task conducted with international raters, which could represent any possible region. We recall that in the IRep annotation task, the raters were able to select multiple expressions for each input video, which means that we deal with a multi-label annotation task. To compute rater agreement on this task, we used Cohen's $\kappa$ implementation, which uses the MASI distance\footnote{NLTK: \url{https://www.nltk.org/_modules/nltk/metrics/distance.html}} \cite{passonneau2006measuring}. In short, MASI is a distance metric used to compare two sets, in our case, two sets of annotated emotions. In Table~\ref{tab:agreement_analysis}, we observe that the repetition in which international raters are used, $R_{4}$, has the lowest agreement across all emotions. When inspecting agreement on individual emotions (see Table~\ref{tab:irr_options_irep} in the Appendix), we observe that for almost all emotions, the IRR scores in $R_4$, the international repetition, have the lowest values. 

Our stability analysis (see Table~\ref{tab:stability_irep} in the Appendix) shows that emotion scores are poorly correlated across repetitions. We observe many weak correlations and only a few moderate correlations,  statistically significant. Furthermore, the correlations with $R_{4}$ seem consistently lower. While the analysis performed by \citet{DBLP:conf/acl/WongPA20} showed that for certain emotions such as ``love'' or ``sadness'' raters can have both high internal agreement and cross-replication agreement when comparing $R\_4$ with the other three repetitions we can not draw such conclusions. Overall, both the internal agreement (see Table \ref{tab:irr_options_irep} in the Appendix) and the cross-replication agreement (see Figure \ref{fig:irep_xrr} in the Appendix) indicate less consistency. More precisely, we can infer that disagreement seems to be intrinsic to the diversity of the raters and the way they interpret emotions.

\subsection{PR: Annotation task and Dataset}
\label{subsec:results_pr}

The inter-rater reliability scores computed on the PR datasets show, overall, fair agreement among raters. To better understand these agreement values, we also computed the IRR scores for each possible option that the raters could have chosen (see Table~\ref{tab:irr_options_pr} in the Appendix), and we observed that the option ``Fit \& Aspect'' is consistently generating lower agreement values, \emph{i.e.}, in each repetition, compared to the other two options. Furthermore, we also observe a considerable difference of 0.21 in IRR scores between repetitions $R_{1}$ and $R_{4}$. As reported by \citet{qarout2019platform}, raters participating in $R_{4}$ had indeed lower accuracy compared to all other repetitions when compared against a ground truth, but the difference does not seem to be significant. 

\begin{figure}[ht!]
\centering
\begin{subfigure}[b]{0.49\columnwidth}
    \centering
    \includegraphics[width=0.99\textwidth]{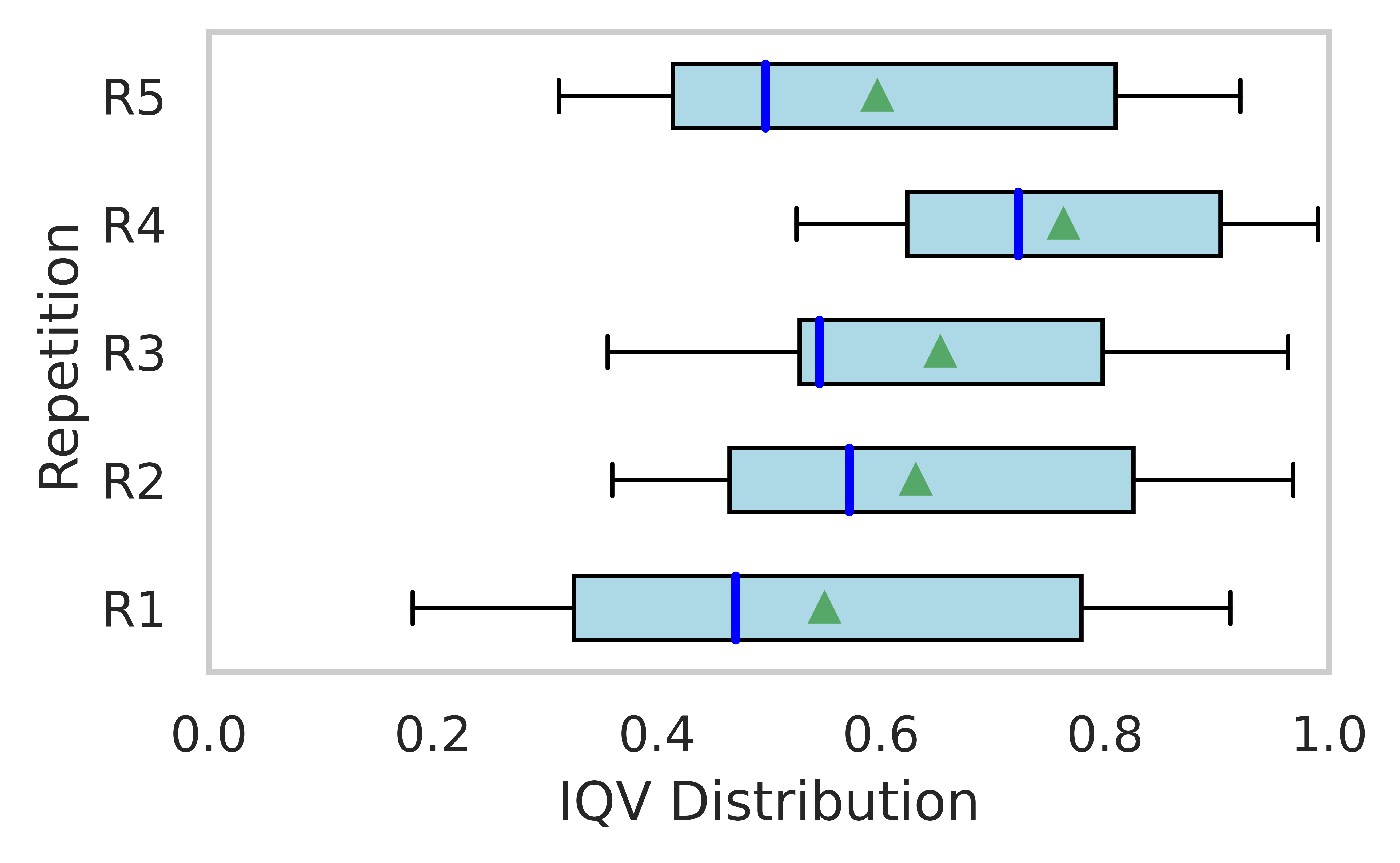}
    \caption{PR dataset}
    \label{fig:pr_iqv}
\end{subfigure} 
\hfill
\begin{subfigure}[b]{0.49\columnwidth}
    \centering
    \includegraphics[width=0.99\textwidth]{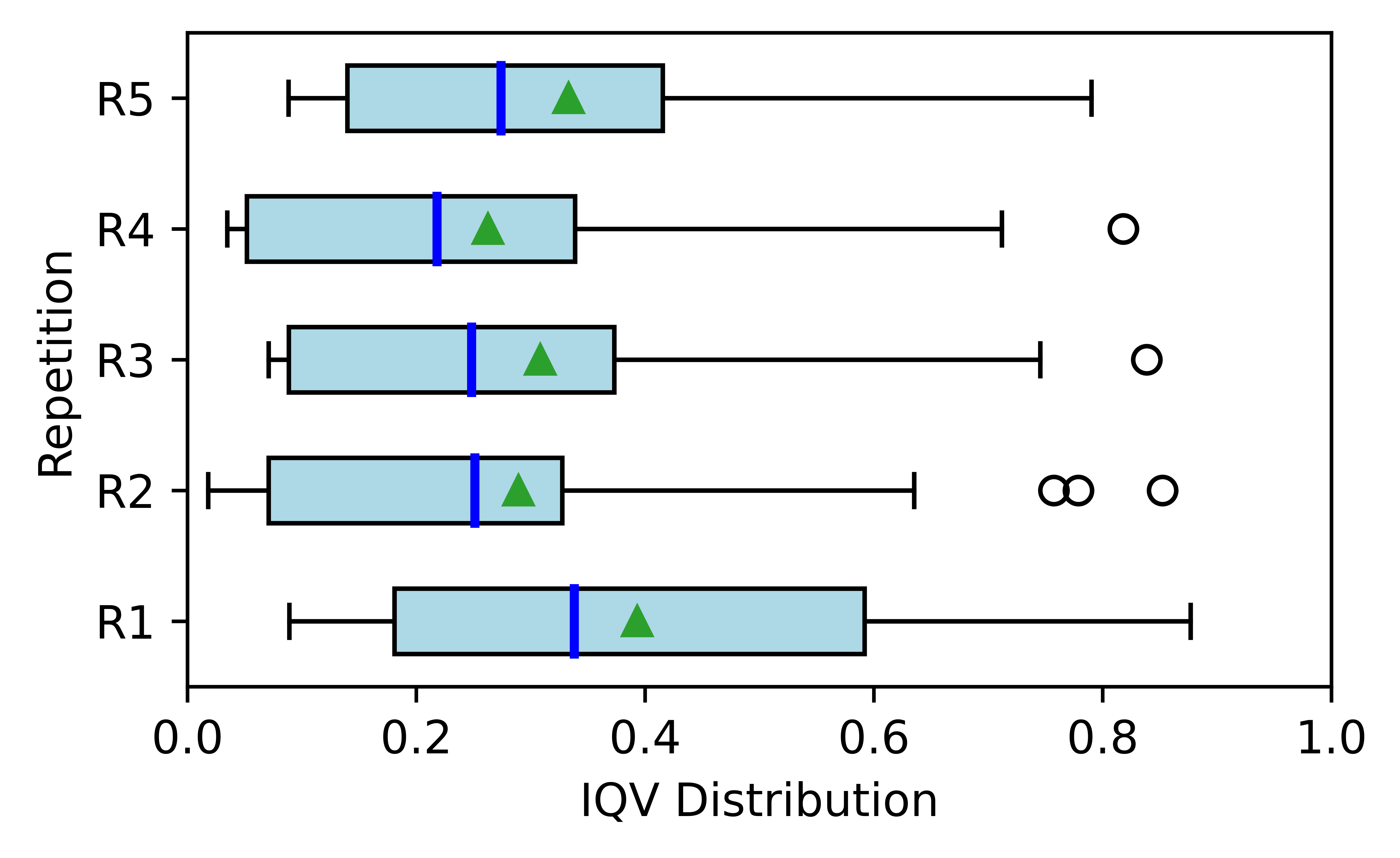}
    \caption{CT dataset. }
    \label{fig:ct_iqv}
\end{subfigure}
\caption{IQV distribution for each unit and repetition ($R_{1}$ to $R_{5}$). For each repetition, the distribution is shown as a boxplot, the blue line represents the median, and the green triangle represents the mean value.}
\end{figure}

For the variability analysis, we computed for each unit in the dataset and, for each repetition, the index of qualitative variation (IQV). In Figure~\ref{fig:pr_iqv}, we observe that in all repetitions, the majority of the units annotated have high variability in terms of categories provided by raters, \emph{i.e.}, it does not seem to be a predominant category that is chosen by the majority of the raters. This, in addition to the low IRR scores for the option ``Fit \& Aspect'', could potentially indicate that the majority of the units annotated in this task are ambiguous or that the options they have to choose from are unclear or have overlapping meanings. The poor agreement and annotations stability across all units in this dataset is also confirmed by our power analysis, which indicates that a very high number of annotations is needed in each repetition to achieve stable results (see Figure~\ref{fig:pr_power_analysis} in the Appendix). In each repetition, the optimal number of raters is around 90, a number that is highly unlikely to be employed in such studies.

Further, we analyzed the stability of the experiments to understand to what extent the results of any two repetitions are similar. For this, we computed for each unit in each repetition the answer given by the majority of the raters (in case of ties, we selected the majority vote at random) and computed the Chi-square test of independence between every two repetitions. On the one hand, this analysis showed that the majority vote answers in any two repetitions are correlated and that there is no statistically significant difference among them (see Table~\ref{tab:pr_stability} in the Appendix). On the other hand, the replicability analysis through the xRR measure showed similarly low values, just as the IRR scores. This result indicates that the agreement among raters is, while low, also consistent across repetitions.

\subsection{CT: Annotation Task and Dataset}
\label{subsec:results_ct}

The raters participating in all repetitions of the crisis tweets (CT) annotation task indicate moderate to substantial agreement, as observed in Table \ref{tab:agreement_analysis}. Similarly, as for the PR datasets, we also computed the IRR scores for each possible option the raters were able to choose. These results are available in Table \ref{tab:irr_options_ct} in the Appendix. For this task, we observe that the majority of options generate moderate to substantial agreement, except for two possible answers, namely ``missing, trapped, or found people'' and ``other useful information''. However, when inspecting the data, we observe that these two options are rarely chosen by raters which might explain their very low agreement values \cite{brenner1996dependence,artstein2008inter}. 

For the variability analysis, we replicated the process described for the PR datasets. In contrast, however, we observe in Figure \ref{fig:ct_iqv} that the index of qualitative evaluation for this dataset has more often values closer to 0, indicating that the annotated units have much lower variability in categories that the raters chose. More precisely, the annotated tweets seem to be easily annotated with a category that is often chosen by the large majority of the raters. $R_{1}$ of the dataset seems to have the lowest precision, which is consistent with the lower IRR score as well as with the lower overall accuracy presented by \citet{qarout2019platform}. In terms of power analysis, similarly to the PR annotation task and dataset, we observe that the mean IRR scores over 100 runs stabilize for a large number of raters (\emph{i.e.}, 85-95 raters).

The stability analysis for this experiment shows that the majority vote answers are very similar across repetitions. More precisely, according to the Chi-square test of independence, we found no statistical difference between the majority vote answers of any two repetitions of the CT task (see Table~\ref{tab:ct_stability} in the Appendix). In terms of cross-rater reliability, the xRR metric shows that the results for this dataset are consistent across repetitions (see Figure~\ref{fig:ct_xrr} in the Appendix). More precisely, we can infer that the high IRR values of these experiments generalize across different rater pools.

\subsection{WS353: Annotation Task and Dataset}
\label{subsec:results_ws}

Among all the replicated studies we analyzed, the highest agreement scores are found for the word similarity datasets, $WS353$, namely 0.59, 0.57, and 0.50, as shown in Table~\ref{tab:agreement_analysis}. Such IRR values are typically acceptable in the context of natural language datasets. As reported by \citet{DBLP:journals/corr/abs-1911-01875}, a thorough precision analysis indicated that while IRR scores have similar values across repetitions, there are certain word pairs for which the similarity score changed dramatically in the second and third repetition (\emph{e.g.}, the pairs ``Maradona''-``football'' and ``Arafat''-``peace'' had higher similarity scores in the first repetition, and very low similarity scores in the second and third repetitions which were run almost 20 years from the first repetition). Our power analysis presented in Figure \ref{fig:ws_power_analysis} in the Appendix indicates that around 12 raters could provide a reliable set of annotations in $R_1$, and even fewer raters in $R_2$ and $R_3$.

\begin{table*}[!ht]
\centering
\caption{Summary of reliability and reproducibility analysis of the nine experimental datasets: we provide a scorecard (see Section~\ref{sec:results}) that further allows to identify influencing factors for data quality and reliability.}
\label{tab:results_summary}
\resizebox{0.75\textwidth}{!}{
\begin{tabular}{lcccccc}
\toprule
\multirow{2}{*}{} & \multicolumn{3}{c}{Reliability} & & \multicolumn{2}{c}{Reproducibility} \\ \cmidrule{2-4} \cmidrule{6-7}
 & \multicolumn{1}{c}{Agreement} & \multicolumn{1}{c}{Variability} & \multicolumn{1}{c}{Power} & & \multicolumn{1}{c}{Stability} & \multicolumn{1}{c}{\begin{tabular}[c]{@{}c@{}}Replicability \\ similarity\end{tabular}} \\ \midrule
VCR\_ALL & low & high & 6 raters & & high & high \\
VCR\_E & low & high & 12 raters & & high & high \\
VCR\_P & low & high & 11 raters & & high & high \\
VCR\_L & low & high & 11 raters & & high & high \\
VCR\_O & low & high & 14 raters & & high & high \\
IRep & low & - & - & & low & low \\
PR & low to medium & high & 90 raters & & high & high \\
CT & medium to high & low & 85-95 raters & & high & high \\
WS353 & medium & low to high & 12 raters & & medium to high & medium to high \\ \bottomrule
\end{tabular}
}
\end{table*}

In terms of stability analysis, the Spearman's $\rho$ correlation shows that all three repetitions are correlated with each other (statistically significant), and in particular $R_{2}$ and $R_{3}$, repetitions that were run on the same platform, with raters having similar characteristics ($R_1$ \& $R_2$: $\rho$=0.87, $p$=9.93e-109; $R_1$ \& $R_2$: $\rho$=0.84, $p$=4.61e-96; $R_2$ \& $R_3$: $\rho$=0.95, $p$=2.88e-182). Similar results are observed in terms of cross-replication reliability, where the xRR values show higher agreement among raters that participated in the last two repetitions (\emph{i.e.}, $R_2$ \& $R_3$: xRR = 0.53), compared to raters that participated in the first repetition and the subsequent ones (\emph{i.e.}, $R_1$ \& $R_2$: xRR=0.49; $R_1$ \& $R_2$: xRR=0.44).

\section{Discussion}
We discuss the results of our methodology for providing a coherent overview of data quality in terms of human factors influencing the reliability and reproducibility of a crowdsourced data collection. Our discussion is driven by the scorecards produced by our proposed methodology for in-depth analysis of the reliability and reproducibility of data annotation studies. The summary of our analysis is presented in Table~\ref{tab:results_summary}. Furthermore, we provide lessons learned for responsible data collection practices, reflect on the limitations of our approach, and give directions for future work.

\paragraph{Factors influencing the quality of data collection.} 
While we surveyed extensive literature in the area of crowdsourcing and human computation, we only found a handful of tasks and datasets that we could identify as repeated experiments. Furthermore, the nine annotation tasks and datasets we identified did not necessarily focus on identifying the factors that could influence the quality of data collection and neither on systematic measurement of their reliability and reproducibility. More precisely, current approaches typically use limited quality and reliability metrics, such as IRR scores or accuracy metrics against a gold standard to gauge data quality. Instead, our metrics provide a scorecard for comparing the reliability and reproducibility of each dataset and surfaces specific factors influencing results' quality. 

In the \textbf{VCR} annotation tasks and datasets, we observed low IRR scores in all repetitions. However, the tasks and datasets have high stability, as the xRR analysis revealed similar cross-rater reliability and highly correlated relevance scores of the video concepts across repetitions. This indicates that raters are similarly consistent within each repetition and across repetitions and that the disagreement indicated by the low IRR scores is, in fact, intrinsic to the subjective nature of the task. One repetition of the \textbf{IRep} annotation task employed international raters. Overall, compared to the other repetitions (\emph{i.e.}, region-specific), our analysis shows consistently low stability and cross-replication reliability. This indicates that not all rater pools are equally consistent across repetition and, more importantly, that the rater disagreement is correlated with the diverse background of the raters influencing the way they interpret emotions. More precisely, for similar tasks, our analysis indicates that \textit{diverse raters should not be expected to produce a coherent view of the annotations} and we advise repeating the data collection by creating dedicated pools of raters with similar demographic characteristics and comparing their results. In the \textbf{PR} annotation task and dataset, we found that while stability can be achieved, the variability analysis and the power analysis indicated that even a very high number of raters (around 90) can exhibit high levels of consistent disagreement typically caused by the subjectivity of the task. In this case, we would advise optimizing the task design in order to decrease additional ambiguity in the annotation categories. The IRR analysis on the individual tweet categories on the \textbf{CT} task indicated that some categories may not be as clear as others or may only seldom be applicable. This indicates that careful attention should again be given to the design, instructions, and possible answer categories in the annotation task. Furthermore, the \textit{high number of raters needed to obtain stable results} indicates that the task might benefit from a more thorough selection of raters and training sessions. Finally, the high variability for certain word pairs in the \textbf{WS353} task indicates that \textit{data collection practices are affected by temporal and familiarity aspects}. This has serious implications for when data collections are reused, as certain annotations may become obsolete or change in interpretation over time. 

\paragraph{Recommendations for responsible data collection}
In sum, applying our proposed methodology for responsible data collection does not pose any requirements on how data is structured or formatted. What we propose, does, however, affect the current practice and assumes a significant adaptation on the use of reliability and reproducibility metrics. The proposed methodology is centered around a set of systematic, iterative (i.e., repeated) pilots which allows to measure different characteristics of the data and task, as well as to capture raters characteristics and measure their potential biases. These aspects are captured with the proposed set of reliability and reproducibility metrics. Finally, we argue for systematic reporting on data collection provenance. 
    
    \textbf{Systematic piloting:} The proposed methodology for guiding data collection with a set of metrics for in-depth, iterative analysis of data reliability and replicability is primarily suitable as an investigative pilot of data annotation studies. Such early experimentation and thorough analysis of annotations would provide specific factors that could influence the data collection and could be ultimately mitigated for large-scale data collection. 

    \textbf{Capture raters, task, and dataset characteristics:} We argue that a responsible data collection practice should borrow guidelines for reporting human-centric studies from the fields of psychology, medicine, and even human-computer interaction, where human stances, opinions, and other meaningful characteristics are thoroughly recorded. While such a process would definitely increase the cost and time to gather the necessary data, it would also allow for more informed decisions on the proper process of collecting raters' annotations and possible future reuse.
    
    \textbf{Cognitive biases assessment:} Recent research has demonstrated that raters' cognitive biases can strongly affect their annotations and reduce data quality~\cite{hube2019understanding,eickhoff2018cognitive,draws2022effects}. To combat the influence of cognitive biases, \citet{draws2021checklist} introduced a checklist that can be used to identify and subsequently measure, mitigate, and document cognitive biases that may present an issue in the data collection tasks. We recommend using such a checklist between each iteration to surface possible cognitive biases that may affect annotations and allow for appropriate mitigation.

    \textbf{Provenance for data collection:} To facilitate responsible reuse of datasets, data documentation, and maintenance approaches should thoroughly record its provenance, including quality scorecards. This would alleviate issues regarding the handling of data, reuse or modifications of annotation tasks, and platform selection. With proper provenance documentation, it is easier to identify factors that could influence data collections' quality. Such requirement becomes clear when data collection is influenced by temporal and regional aspects (see WS353 and IRep).

\subsection{Limitations and Future Work}
\label{subsec:limitations}

\textbf{Diversity and scale of datasets.} We experimented with nine datasets and annotation tasks with various goals, modalities, sizes, and overall setups. We repeated each task three to five times. While the overall number of units in some datasets was small (\emph{e.g.}, $\sim20$ input units), the overall dataset size was much bigger as the number of raters providing annotations per item was significantly larger than in usual data collections. Our methodology is agnostic to the dataset size, and the significance of the results is not influenced by dataset size. In future work, we could extend the analysis to other data annotation tasks and input data modalities. 

\textbf{Scale and optimal number of repetitions.} The repeatability experiments may not be scalable in terms of time and cost. However, our methodology provides optimization criteria that can mitigate this limitation in terms of input for the annotation tasks and the use of the bootstrap technique to optimize the number of raters needed for reliable results. As we have shown, iterative instances of \emph{collect}, \emph{measure}, \emph{repeat} are suitable for adhering to responsible data collection practices. However, it is not trivial to decide on the number of repetitions necessary to determine that the collected data is reliable. This can be even more problematic for more subjective annotation tasks, which can be affected by raters' interpretations, opinions, perspectives, or familiarity with the items. Future work can address this limitation by investigating additional metrics for determining the suitable number of repetitions. Furthermore, future work could also focus on proposing a single suitable reproducibility score for repeated experiments similar to the one proposed by \citet{DBLP:conf/acl/BelzPM22} for system reproducibility.

\textbf{Raters' characteristics.} The analyzed tasks included only limited information about the raters, besides some very general characteristics such as the platform on which they were recruited, country, or HIT approval rates. Furthermore, only one dataset out of the nine analyzed had a substantially different population of raters (i.e., from different countries). This aspect limits our analysis in terms of additional human factors that could influence rater agreement and the stability of the collected annotations. Future work could focus on replicating our analysis on more controlled data annotation experiments to study, for instance, the impact of age, gender, and other demographic information as additional reliability factors for responsible data collection. In future work, this iterative method of addressing responsible data collection should also investigate ways of properly maintaining and describing data provenance. 

\section{Conclusions}
The continuous deployment of AI systems powered by crowdsourced data in real-world tasks has increased the attention of the research community to further scrutinize the quality and reliability of such datasets in diverse settings. In this paper, we propose a Responsible AI (RAI) methodology designed to guide the data collection with a set of metrics for an in-depth, iterative analysis of the \emph{human factors influencing the quality and reliability} of the data they generate. The methodology consists of a set of metrics for a systematic analysis of data that brings transparency in how to interpret human disagreement and how to validate rater quality assuming diverse settings. We propose an iterative process to measure the reliability and stability of crowdsourced data from different perspectives. The repetition of experiments allows us to perform a comparative analysis across repetitions and measure changes both in the annotations and in the variance and consistency of raters. Due to the variety of quality metrics we employ, this research can have a strong impact on the way we measure data quality based on subjective human ratings. This further leads to increased diversity of AI systems and helps us deal with fairness and accountability aspects in data collection. By making the analysis process transparent through the set of metrics, we also deal with fairness and accountability aspects in data collection. We validated our methodology on nine existing annotation tasks and datasets. We found that our systematic set of metrics allows us to draw insights into the human and task-dependent factors that influence the quality of AI datasets.

\bibliography{bibfile}

\appendix

\section{Agreement Analysis}

\setcounter{table}{0}
\renewcommand{\thetable}{A\arabic{table}}

We report here on the agreement analysis performed on the product review (PR) dataset (Table~\ref{tab:irr_options_pr}), crisis tweets (CT) dataset (Table~\ref{tab:irr_options_ct}), and video human facial expressions (IRep) dataset (Table~\ref{tab:irr_options_irep}).

\begin{table}[!ht]
\centering
\caption{Inter-rater reliability scores on each category the raters were able to choose in the PR annotation task. The IRR score is computed using the Krippendorff's $\alpha$ metric.}
\label{tab:irr_options_pr}
\resizebox{1\columnwidth}{!}{
\begin{tabular}{llllll} \toprule

& R1   & R2   & R3   & R4   & R5   \\ \midrule
Fit \& Aspect & 0.24 & 0.17 & 0.20 & 0.11 & 0.22 \\
Size \& Aspect & 0.44 & 0.37 & 0.35 & 0.24 & 0.41 \\
No Issue with Size \& Aspect & 0.51 & 0.43 & 0.39 & 0.25 & 0.43 \\ \bottomrule
\end{tabular}
}
\end{table}

\begin{table}[!h]
\centering
\caption{Inter-rater reliability scores on each category the raters were able to choose in the CT annotation task. The IRR score is computed using the Krippendorff's $\alpha$ metric.}
\label{tab:irr_options_ct} 
\resizebox{1\columnwidth}{!}{
\begin{tabular}{llllll} \toprule
 & R1    & R2   & R3     & R4     & R5    \\ \midrule
Infrastructure and utilities damage               & 0.44  & 0.56 & 0.53   & 0.57   & 0.52  \\
Injured or dead people                            & 0.82  & 0.93 & 0.90   & 0.95   & 0.88  \\
Caution and advice                                & 0.42  & 0.56 & 0.56   & 0.51   & 0.52  \\
Donation needs or offers or volunteering & 0.78  & 0.83 & 0.84   & 0.87   & 0.81  \\
Sympathy and emotional support                    & 0.70  & 0.80 & 0.81   & 0.86   & 0.73  \\
Displaced people and evacuations                  & 0.50  & 0.72 & 0.72   & 0.80   & 0.68  \\
Missing, trapped, or found people                 & 0.003 & 0.01 & -0.003 & -0.001 & 0.003 \\
Not related or irrelevant                         & 0.55  & 0.59 & 0.57   & 0.67   & 0.57  \\
Other useful information                          & 0.22  & 0.26 & 0.24   & 0.29   & 0.29 \\ \bottomrule
\end{tabular}
}
\end{table}

\begin{table}[!h]
\centering
\caption{Inter-rater reliability scores on each category the raters were able to choose in the IRep annotation task. The IRR score is computed using the Krippendorff's $\alpha$ metric.}
\label{tab:irr_options_irep}
\resizebox{0.95\columnwidth}{!}{
\begin{tabular}{lllll} \toprule
               & R1     & R2     & R3     & R4    \\ \midrule
Amusement      & 0.42   & 0.12   & 0.66   & 0.27  \\
Anger          & 0.45   & 0.39   & 0.72   & 0.44  \\
Awe            & 0.25   & 0.19   & 0.25   & 0.09  \\
Boredom        & 0.49   & 0.38   & 0.58   & 0.30  \\
Concentration  & 0.34   & 0.40   & 0.56   & 0.21  \\
Confusion      & 0.23   & 0.22   & 0.41   & 0.13  \\
Contemplation  & 0.13   & 0.09   & 0.64   & 0.10  \\
Contempt       & 0.16   & 0.24   & 0.53   & 0.11  \\
Contentment    & 0.23   & 0.50   & 0.71   & 0.10  \\
Desire         & 0.54   & 0.69   & 0.83   & 0.14  \\
Disappointment & 0.19   & 0.18   & 0.62   & 0.16  \\
Disgust        & 0.52   & 0.20   & 0.50   & 0.14  \\
Distress       & 0.08   & 0.10   & 0.26   & 0.13  \\
Doubt          & 0.19   & 0.18   & 0.22   & -0.02 \\
Ecstasy        & 0.13   & 0.24   & 0.50   & 0.17  \\
Elation        & 0.12   & 0.21   & 0.66   & 0.15  \\ 
Embarrassment  & 0.19   & -0.007 & 0.28   & 0.06  \\
Error.other    & 1.0    & 1.0    & 1.0    & 0.28  \\
Fear           & 0.43   & 0.36   & 0.67   & 0.31  \\
Interest       & 0.22   & 0.08   & 0.44   & 0.13  \\
Love           & 0.66   & 0.61   & 0.72   & 0.32  \\
Neutral        & 0.41   & 0.15   & 0.50   & 0.13  \\
Pain           & 0.15   & 0.44   & 0.46   & 0.13  \\
Pride          & 0.17   & 0.07   & 0.39   & 0.17  \\
Realization    & 0.09   & -0.003 & 0.33   & 0.03  \\
Relief         & 0.17   & 0.28   & 0.53   & 0.05  \\
Sadness        & 0.51   & 0.44   & 0.54   & 0.22  \\
Shame          & -0.006 & -0.002 & 0.0    & 0.17  \\
Surprise       & 0.35   & 0.44   & 0.67   & 0.20  \\
Sympathy       & 0.19   & 0.07   & 0.51   & -0.02 \\
Triumph        & 0.42   & 0.22   & 0.28   & 0.08  \\
Unsure         & 0.61   & 0.55   & -0.002 & 1.0  \\ \bottomrule
\end{tabular}
}
\end{table}


\section{Variability Analysis}

\setcounter{table}{0}
\renewcommand{\thetable}{B\arabic{table}}

Table~\ref{tab:reliability_individual_replication} shows an overview of the raters variability in each repetition of the video concept relevance (VCR) tasks. For each video - concept pair, we computed the standard deviation of their score. The majority of the experiments have a mean standard deviation (MSTD) of around 0.3, with the task VCR\_O having a higher value, of around 0.36. The standard deviation of deviations (STDD) is similar across tasks and repetitions, with the lowest value observed for the VCR\_O task. These high values observed for MSTD and STDD show that video concepts relevance annotation is subjective, with raters often disagreeing. Concepts of type organization seem to generate the most disagreement among annotators.

\begin{table*}[!h]
\centering
\caption{Overview of rater variability metrics for all \emph{VCR} annotation tasks and their repetitions ($R_1$, $R_2$ and $R_3$).}
\label{tab:reliability_individual_replication}
\resizebox{0.99\textwidth}{!}{
\begin{tabular}{cccccccccccccccccccc} \toprule
\multirow{2}{*}{Measure} & \multicolumn{3}{c}{VCR\_ALL} & & \multicolumn{3}{c}{VCR\_E}  &   & \multicolumn{3}{c}{VCR\_P} & & \multicolumn{3}{c}{VCR\_L} & & \multicolumn{3}{c}{VCR\_O} \\ \cmidrule{2-4} \cmidrule{6-8} \cmidrule{10-12} \cmidrule{14-16} \cmidrule{18-20}
 & $R_1$ & $R_2$ & $R_3$ & & $R_1$ & $R_2$ & $R_3$ &   & $R_1$ & $R_2$ & $R_3$ & & $R_1$ & $R_2$ & $R_3$ & & $R_1$ & $R_2$ & $R_3$\\ \midrule
Mean Stdev (MSTD) & 0.28 & 0.31 & 0.30 & & 0.24 & 0.28 & 0.28 & & 0.31 & 0.31 & 0.32 & & 0.31 & 0.32 & 0.31 & & 0.36 & 0.37 & 0.35 \\ \midrule
Stdev of Deviation (STDD) & 0.20 & 0.20 & 0.19 & & 0.22 & 0.20 & 0.21 & & 0.19 & 0.20 & 0.19 & & 0.20 & 0.19  & 0.20 & & 0.17 &   0.16 & 0.15    \\ \bottomrule
\end{tabular}
}
\end{table*}

\setcounter{figure}{0}
\renewcommand{\thefigure}{C\arabic{figure}}

\begin{figure*}[h!]
  \begin{subfigure}[b]{0.99\textwidth}
    \includegraphics[width=\textwidth]{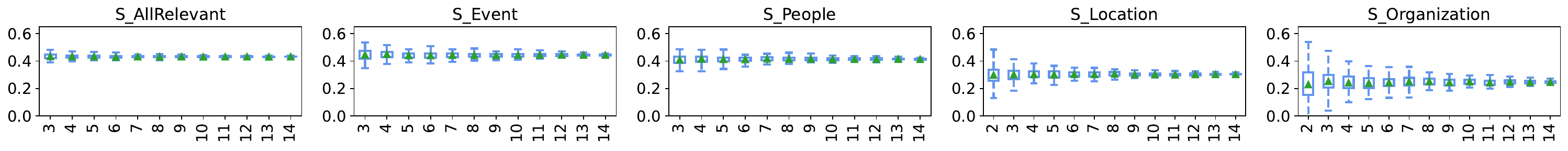}
    \caption{$R_1$}
    \label{fig:bootstrap_workers_R1}
  \end{subfigure}
  \begin{subfigure}[b]{0.99\textwidth}
    \includegraphics[width=\textwidth]{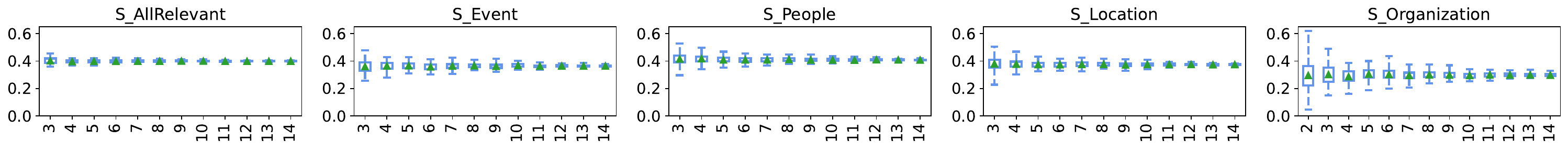}
    \caption{$R_2$}
    \label{fig:bootstrap_workers_R2}
  \end{subfigure}
  \begin{subfigure}[b]{0.99\textwidth}
    \includegraphics[width=\textwidth]{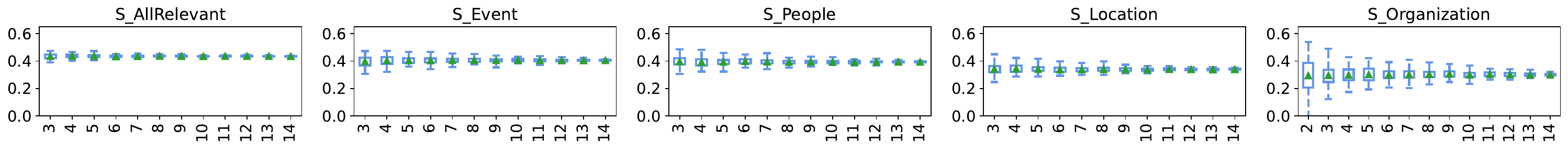}
    \caption{$R_3$}
    \label{fig:bootstrap_workers_R3}
  \end{subfigure}
\caption{Bootstrap the number of raters in each repetition ($R_1$ to $R_3$) of the VCR annotation tasks. The plot shows the distribution of the IRR scores per each number of raters, where the number of raters ranges from 0 to 15. The green triangle represents the mean value.}
\label{fig:bootstrap_workers}
\end{figure*}

\begin{figure*}[ht!]
    \centering
    \includegraphics[width=0.99\textwidth]{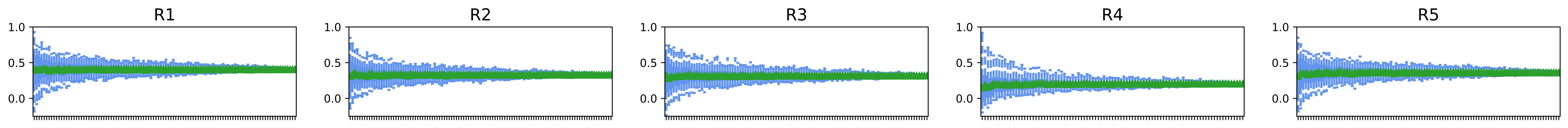}
    \caption{Bootstrap the number of raters in each repetition ($R_1$ to $R_5$) of the PR annotation task. The plot shows the distribution of the IRR scores per each number of raters, where the number of raters ranges from 0 to the maximum number of raters employed in the given repetition. The green triangle represents the mean value.}
    \label{fig:pr_power_analysis}
\end{figure*}

\begin{figure*}[h!]
    \centering
    \includegraphics[width=0.99\textwidth]{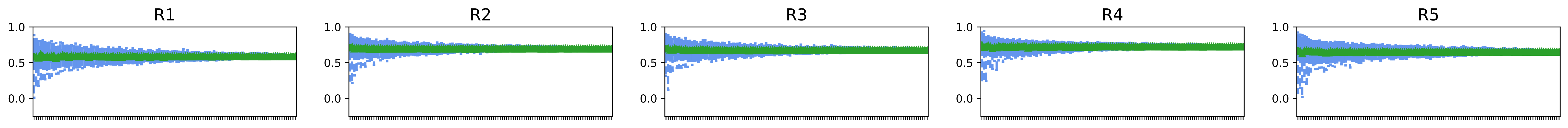}
    \caption{Bootstrap the number of raters in each repetition ($R_1$ to $R_5$) of the CT annotation task. The plot shows the distribution of the IRR scores per each number of raters, where the number of raters ranges from 0 to the maximum number of raters employed in the given repetition. The green triangle represents the mean value.}
    \label{fig:ct_power_analysis}
\end{figure*}


\section{Power analysis}

\begin{figure*}[h!]
    \centering
    \includegraphics[width=0.99\textwidth]{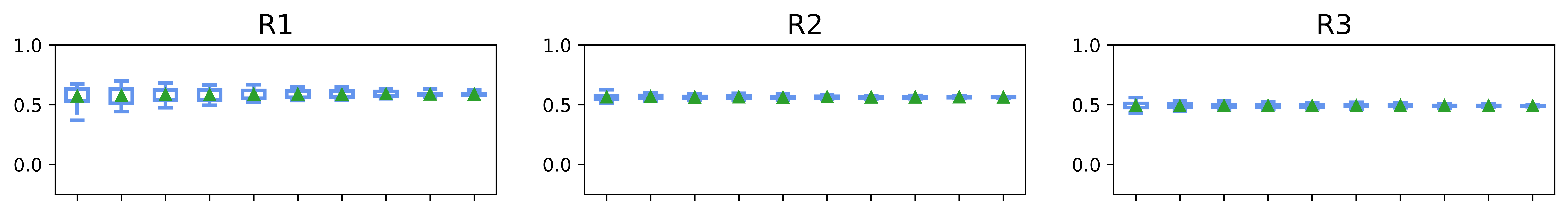}
    \caption{Bootstrap the number of raters in each repetition ($R_1$ to $R_3$) of the WS353 annotation task. The plot shows the distribution of the IRR scores per each number of raters, where the number of raters ranges from 0 to the maximum number of raters employed in the given repetition. The green triangle represents the mean value.}
    \label{fig:ws_power_analysis}
\end{figure*}

In this section, we show additional results on the power analysis performed on the following tasks and datasets: video concept relevance (VCR) in Figure~\ref{fig:bootstrap_workers}, product reviews (PR) in Figure~\ref{fig:pr_power_analysis}, crisis tweets (CT) in Figure~\ref{fig:ct_power_analysis}, and word similarity (WS353) in Figure~\ref{fig:ws_power_analysis}. We recall here that the power analysis is performed using bootstrap experiments on the number of raters. This allows us to observe the impact of the number of raters per unit (in each task and repetition) on the inter-rater reliability score, computed in our case using Krippendorff's $\alpha$. Thus, in the aforementioned figures, we show the IRR distribution when we bootstrap for each number of raters 100 times, on every task and repetition. We observe that all repetitions of all tasks tend to display similar variability in terms of IRR scores.

\setcounter{table}{0}
\renewcommand{\thetable}{D\arabic{table}}

\begin{table*}[!ht]
\centering
\caption{Spearman's $\rho$ rank correlation of the relevance of each emotion in the IRep annotation task and datasets.}
\resizebox{1.0\textwidth}{!}{
\label{tab:stability_irep} 
\begin{tabular}{lcccccc} \toprule
& $R\_{1}$ \& $R\_{2}$ & $R\_{1}$ \& $R\_{3}$ & $R\_{1}$ \& $R\_{4}$ & $R\_{2}$ \& $R\_{3}$ & $R\_{2}$ \& $R\_{4}$ & $R\_{3}$ \& $R\_{4}$  \\ \midrule
amusement
& $\rho$=0.45, $p$=2.13e-55
& $\rho$=0.60, $p$=5.27e-106
& $\rho$=0.43, $p$=6.54e-49
& $\rho$=0.38, $p$=3.94e-37
& $\rho$=0.39, $p$=6.26e-40
& $\rho$=0.42, $p$=1.71e-46 \\
anger
& $\rho$=0.49, $p$=2.40e-65
& $\rho$=0.53, $p$=6.77e-80
& $\rho$=0.44, $p$=7.31e-51
& $\rho$=0.52, $p$=3.16e-74
& $\rho$=0.50, $p$=5.88e-69
& $\rho$=0.55, $p$=4.25e-84 \\
awe
& $\rho$=0.20, $p$=6.65e-11
& $\rho$=0.10, $p$=0.001
& $\rho$=0.07, $p$=0.02
& $\rho$=0.15, $p$=5.72e-07
& $\rho$=0.07, $p$=0.02
& $\rho$=0.13, $p$=2.08e-05 \\
boredom
& $\rho$=0.34, $p$=9.15e-31
& $\rho$=0.50, $p$=5.71e-70
& $\rho$=0.34, $p$=1.54e-29
& $\rho$=0.45, $p$=1.47e-54
& $\rho$=0.31, $p$=7.30e-25
& $\rho$=0.34, $p$=2.01e-30 \\
concentration
& $\rho$=0.37, $p$=1.23e-35
& $\rho$=0.36, $p$=2.94e-33
& $\rho$=0.30, $p$=6.93e-23
& $\rho$=0.51, $p$=7.76e-71
& $\rho$=0.41, $p$=1.77e-45
& $\rho$=0.39, $p$=2.07e-40 \\
confusion
& $\rho$=0.30, $p$=2.55e-23
& $\rho$=0.27, $p$=1.91e-19
& $\rho$=0.25, $p$=2.17e-16
& $\rho$=0.37, $p$=2.00e-36
& $\rho$=0.20, $p$=1.10e-10
& $\rho$=0.32, $p$=2.92e-26 \\
contemplation
& $\rho$=0.15, $p$=1.62e-06
& $\rho$=0.08, $p$=0.006
& $\rho$=0.14, $p$=6.79e-06
& $\rho$=0.05, $p$=0.11
& $\rho$=0.10, $p$=0.001
& $\rho$=0.15, $p$=6.38e-07 \\
contempt
& $\rho$=0.14, $p$=3.75e-06
& $\rho$=0.24, $p$=7.69e-15
& $\rho$=0.23, $p$=7.75e-14
& $\rho$=0.28, $p$=7.41e-21
& $\rho$=0.13, $p$=2.48e-05
& $\rho$=0.16, $p$=3.52e-07 \\
contentment
& $\rho$=0.43, $p$=6.06e-50
& $\rho$=-0.02, $p$=0.48
& $\rho$=0.13, $p$=2.91e-05
& $\rho$=-0.09, $p$=0.002
& $\rho$=0.13, $p$=3.32e-05
& $\rho$=0.19, $p$=4.29e-10 \\
desire
& $\rho$=0.51, $p$=7.18e-72
& $\rho$=0.51, $p$=1.37e-72
& $\rho$=0.28, $p$=3.48e-20
& $\rho$=0.63, $p$=1.76e-120
& $\rho$=0.25, $p$=3.77e-17
& $\rho$=0.31, $p$=8.54e-25 \\
disappointment
& $\rho$=0.16, $p$=7.40e-08
& $\rho$=0.23, $p$=1.12e-14
& $\rho$=0.14, $p$=8.11e-06
& $\rho$=0.30, $p$=5.40e-24
& $\rho$=0.12, $p$=0.0001
& $\rho$=0.17, $p$=9.18e-09 \\
disgust
& $\rho$=0.26, $p$=1.03e-17
& $\rho$=0.28, $p$=9.20e-21
& $\rho$=0.22, $p$=3.81e-13
& $\rho$=0.30, $p$=2.45e-23
& $\rho$=0.23, $p$=7.11e-14
& $\rho$=0.21, $p$=1.40e-12 \\
distress
& $\rho$=0.19, $p$=2.05e-10
& $\rho$=0.22, $p$=3.27e-13
& $\rho$=0.17, $p$=3.01e-08
& $\rho$=0.17, $p$=1.16e-08
& $\rho$=0.19, $p$=6.30e-10
& $\rho$=0.20, $p$=1.24e-10 \\
doubt
& $\rho$=0.18, $p$=2.42e-09
& $\rho$=0.02, $p$=0.46
& $\rho$=0.02, $p$=0.51
& $\rho$=0.09, $p$=0.004
& $\rho$=0.04, $p$=0.23
& $\rho$=0.01, $p$=0.68 \\
ecstasy
& $\rho$=0.22, $p$=2.67e-13
& $\rho$=0.08, $p$=0.01
& $\rho$=0.13, $p$=1.75e-05
& $\rho$=-0.009, $p$=0.78
& $\rho$=0.15, $p$=1.82e-06
& $\rho$=-0.01, $p$=0.64 \\
elation
& $\rho$=0.29, $p$=1.66e-21
& $\rho$=0.36, $p$=1.27e-33
& $\rho$=0.18, $p$=2.04e-09
& $\rho$=0.40, $p$=9.74e-42
& $\rho$=0.36, $p$=3.78e-34
& $\rho$=0.26, $p$=1.99e-17 \\
embarrassment
& $\rho$=0.14, $p$=3.61e-06
& $\rho$=0.21, $p$=6.35e-12
& $\rho$=0.12, $p$=5.34e-05
& $\rho$=0.21, $p$=9.86e-12
& $\rho$=0.12, $p$=0.0001
& $\rho$=0.08, $p$=0.01 \\
fear
& $\rho$=0.50, $p$=8.38e-70
& $\rho$=0.53, $p$=1.37e-78
& $\rho$=0.38, $p$=3.87e-38
& $\rho$=0.70, $p$=4.58e-156
& $\rho$=0.46, $p$=1.16e-55
& $\rho$=0.48, $p$=6.67e-64 \\
interest
& $\rho$=0.26, $p$=3.87e-18
& $\rho$=0.24, $p$=2.92e-15
& $\rho$=0.17, $p$=9.02e-09
& $\rho$=0.13, $p$=1.34e-05
& $\rho$=0.17, $p$=2.64e-08
& $\rho$=0.15, $p$=4.44e-07 \\
love
& $\rho$=0.61, $p$=2.62e-110
& $\rho$=0.57, $p$=7.00e-93
& $\rho$=0.42, $p$=1.27e-45
& $\rho$=0.72, $p$=6.70e-172
& $\rho$=0.57, $p$=4.42e-94
& $\rho$=0.57, $p$=4.97e-93 \\
neutral
& $\rho$=0.27, $p$=2.47e-19
& $\rho$=0.16, $p$=0.0001
& $\rho$=0.23, $p$=6.46e-14
& $\rho$=0.09, $p$=0.002
& $\rho$=0.22, $p$=1.00e-12
& $\rho$=0.09, $p$=0.004 \\
pain
& $\rho$=0.43, $p$=6.23e-50
& $\rho$=0.27, $p$=7.23e-19
& $\rho$=0.29, $p$=6.68e-22
& $\rho$=0.47, $p$=6.94e-61
& $\rho$=0.29, $p$=5.90e-22
& $\rho$=0.24, $p$=2.34e-15 \\
pride
& $\rho$=0.14, $p$=2.09e-06
& $\rho$=0.29, $p$=7.77e-22
& $\rho$=0.17, $p$=1.47e-08
& $\rho$=0.45, $p$=1.73e-54
& $\rho$=0.16, $p$=1.56e-07
& $\rho$=0.16, $p$=1.05e-07 \\
realization
& $\rho$=-0.02, $p$=0.46
& $\rho$=0.03, $p$=0.26
& $\rho$=0.00004, $p$=0.99
& $\rho$=-0.006, $p$=0.86
& $\rho$=-0.03, $p$=0.37
& $\rho$=0.07, $p$=0.03 \\
relief
& $\rho$=0.09, $p$=0.003
& $\rho$=0.11, $p$=0.0003
& $\rho$=0.12, $p$=0.0001
& $\rho$=0.25, $p$=1.28e-16
& $\rho$=0.10, $p$=0.001
& $\rho$=0.17, $p$=3.96e-08 \\
sadness
& $\rho$=0.51, $p$=1.42e-72
& $\rho$=0.57, $p$=2.14e-93
& $\rho$=0.40, $p$=7.26e-43
& $\rho$=0.52, $p$=3.07e-75
& $\rho$=0.43, $p$=3.25e-50
& $\rho$=0.47, $p$=3.01e-58 \\
shame
& $\rho$=0.33, $p$=3.11e-29
& $\rho$=0.28, $p$=4.83e-20
& $\rho$=0.16, $p$=1.004e-07
& $\rho$=-0.002, $p$=0.94
& $\rho$=-0.01, $p$=0.64
& $\rho$=0.16, $p$=1.91e-07 \\
surprise
& $\rho$=0.49, $p$=6.36e-66
& $\rho$=0.55, $p$=7.92e-87
& $\rho$=0.35, $p$=1.91e-31
& $\rho$=0.55, $p$=2.84e-86
& $\rho$=0.30, $p$=3.35e-24
& $\rho$=0.38, $p$=6.81e-39 \\
sympathy
& $\rho$=0.08, $p$=0.01
& $\rho$=0.14, $p$=2.72e-06
& $\rho$=0.03, $p$=0.35
& $\rho$=0.26, $p$=3.31e-18
& $\rho$=0.12, $p$=5.03e-05
& $\rho$=0.15, $p$=1.37e-06 \\
triumph
& $\rho$=0.44, $p$=6.66e-53
& $\rho$=0.53, $p$=2.21e-74
& $\rho$=0.26, $p$=2.05e-18
& $\rho$=0.65, $p$=5.37e-127
& $\rho$=0.22, $p$=4.55e-13
& $\rho$=0.30, $p$=1.01e-23 \\
unsure
& $\rho$=0.66, $p$=5.72e-135
& $\rho$=0.16, $p$=2.12e-07
&
& $\rho$=0.11, $p$=0.0002 
& 
&\\ \bottomrule
\end{tabular}
}
\end{table*}

\begin{table*}[!ht]
\centering
\caption{Stability analysis on the PR annotation task and dataset. The table indicates the correlation of the aggregated raters’ annotations (\emph{i.e.}, using majority vote) between pairwise repetitions of the PR task.}
\label{tab:pr_stability}
\resizebox{0.9\textwidth}{!}{
\begin{tabular}{lllll} \toprule
   & \multicolumn{1}{c}{R2}      & \multicolumn{1}{c}{R3}      & \multicolumn{1}{c}{R4}      & \multicolumn{1}{c}{R5}       \\ \midrule
R1 & $\chi$ = 24.87, $p$ = 5.35e-05 & $\chi$ = 35.0, $p$ = 4.65e-07  & $\chi$ = 34.55, $p$ = 5.76e-07 & $\chi$ = 29.80, $p$ = 5.38e-06  \\
R2 &                             & $\chi$ = 29.88, $p$ = 5.19e-06 & $\chi$ = 29.70, $p$ = 5.64e-06 & $\chi$ = 24.87, $p$ = 5.35e-05 \\
R3 &                             &                             & $\chi$ = 30.91, $p$ = 3.19e-06 & $\chi$ = 35.0, $p$ = 4.65e-07   \\
R4 &                             &                             &                             & $\chi$ = 25.68, $p$ = 3.67e-05 \\ \bottomrule
\end{tabular}
}
\end{table*}

\begin{table*}[!ht]
\centering
\caption{Stability analysis on the CT annotation task and dataset. The table indicates the correlation of the aggregated raters’ annotations (\emph{i.e.}, using majority vote) between pairwise repetitions of the CT task.}
\label{tab:ct_stability}
\resizebox{0.9\textwidth}{!}{
\begin{tabular}{cllll} \toprule
   & \multicolumn{1}{c}{R2}      & \multicolumn{1}{c}{R3}      & \multicolumn{1}{c}{R4}      & \multicolumn{1}{c}{R5}       \\ \midrule
R1 & $\chi$ = 23.63, $p$ = 9.48e-05 & $\chi$ = 19.56, $p$ = 0.0006  & $\chi$ = 23.63, $p$ = 9.48e-05 & $\chi$ = 24.04, $p$ = 7.81e-05  \\
R2 &                             & $\chi$ = 35.58, $p$ = 3.52e-07 & $\chi$ = 40.70, $p$ = 4.33e-08 & $\chi$ = 27.94, $p$ = 1.28e-05 \\
R3 &                             &                             & $\chi$ = 35.58, $p$ = 3.52e-07 & $\chi$ = 32.22, $p$ = 1.72e-06   \\
R4 &                             &                             &                             & $\chi$ = 27.94, $p$ = 1.28e-05 \\ \bottomrule
\end{tabular}
}
\end{table*}

\section{Stability analysis}

The stability analysis is performed by measuring the correlation between pairwise repetitions of a task. For the tasks at hand, we compute the correlations using the Spearman's rank correlation for the IRep annotation task and datasets (in Table~\ref{tab:stability_irep}) and using the Chi-square test of independence for the PR (in Table~\ref{tab:pr_stability}) and CT (in Table~\ref{tab:ct_stability}) annotation tasks and datasets.

\section{Replicability similarity analysis}

\setcounter{figure}{0}
\renewcommand{\thefigure}{E\arabic{figure}}

\begin{figure}[ht!]
    \centering
    \includegraphics[width=0.32\textwidth]{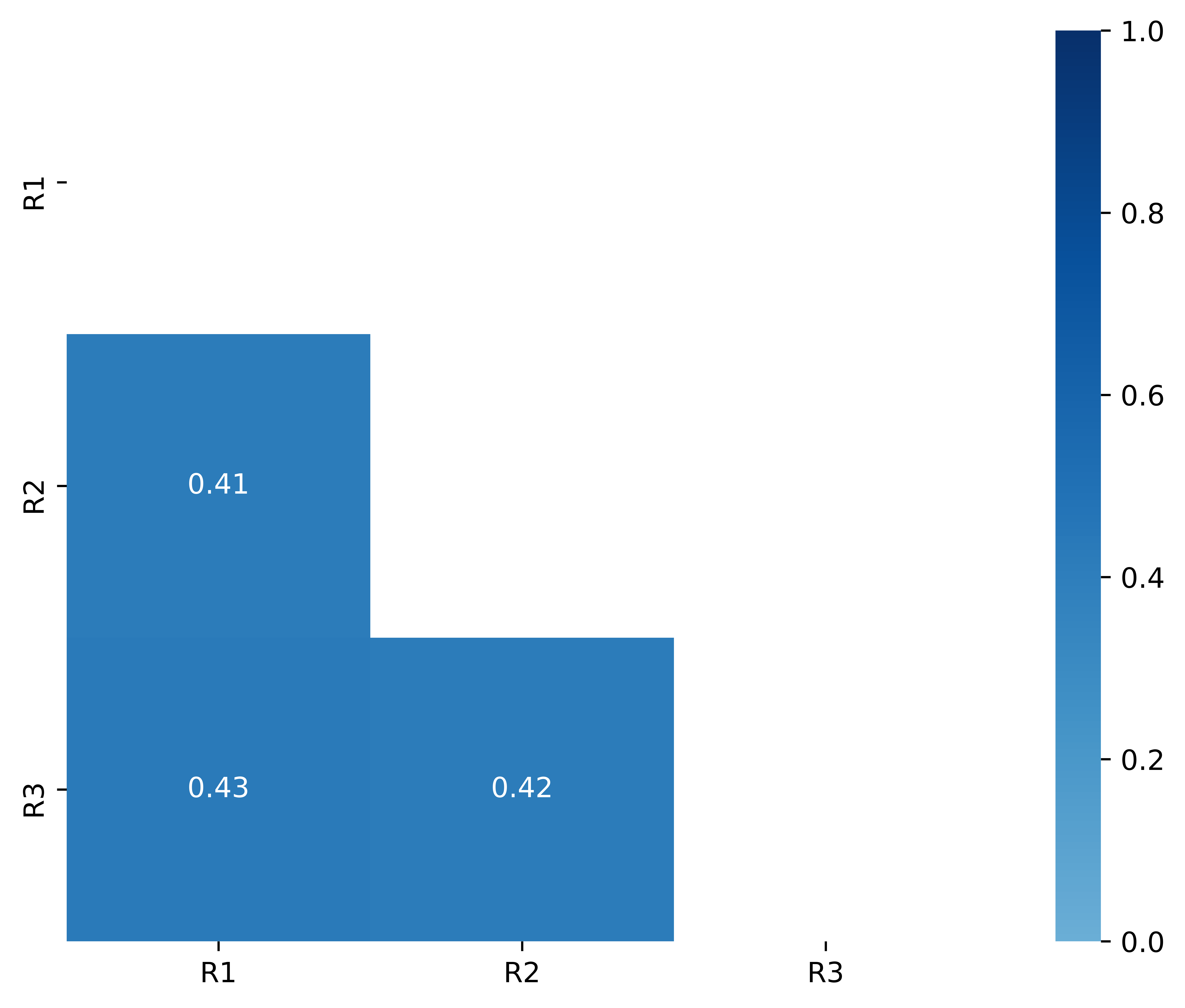}
    \caption{Cross-rater reliability analysis on the VCR\_ALL annotation task and datasets. Each cell in the heatmap represents the xRR score between pairwise sets of repetitions.}
    \label{fig:vcr_all_xrr}
\end{figure}

\begin{figure}[ht!]
    \centering
    \includegraphics[width=0.32\textwidth]{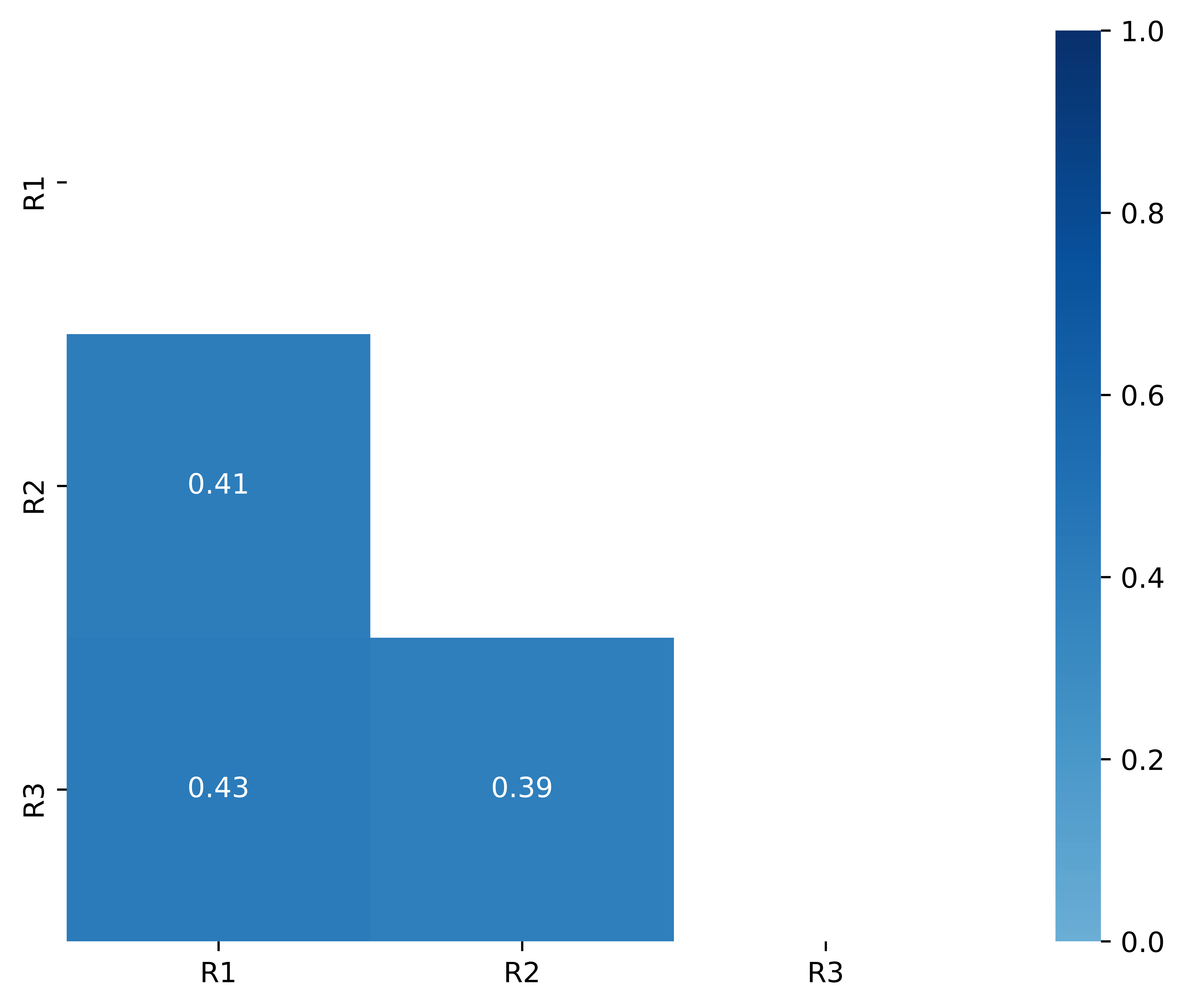}
    \caption{Cross-rater reliability analysis on the VCR\_E annotation task and datasets. Each cell in the heatmap represents the xRR score between pairwise sets of repetitions.}
    \label{fig:vcr_e_xrr}
\end{figure}

\begin{figure}[ht!]
    \centering
    \includegraphics[width=0.32\textwidth]{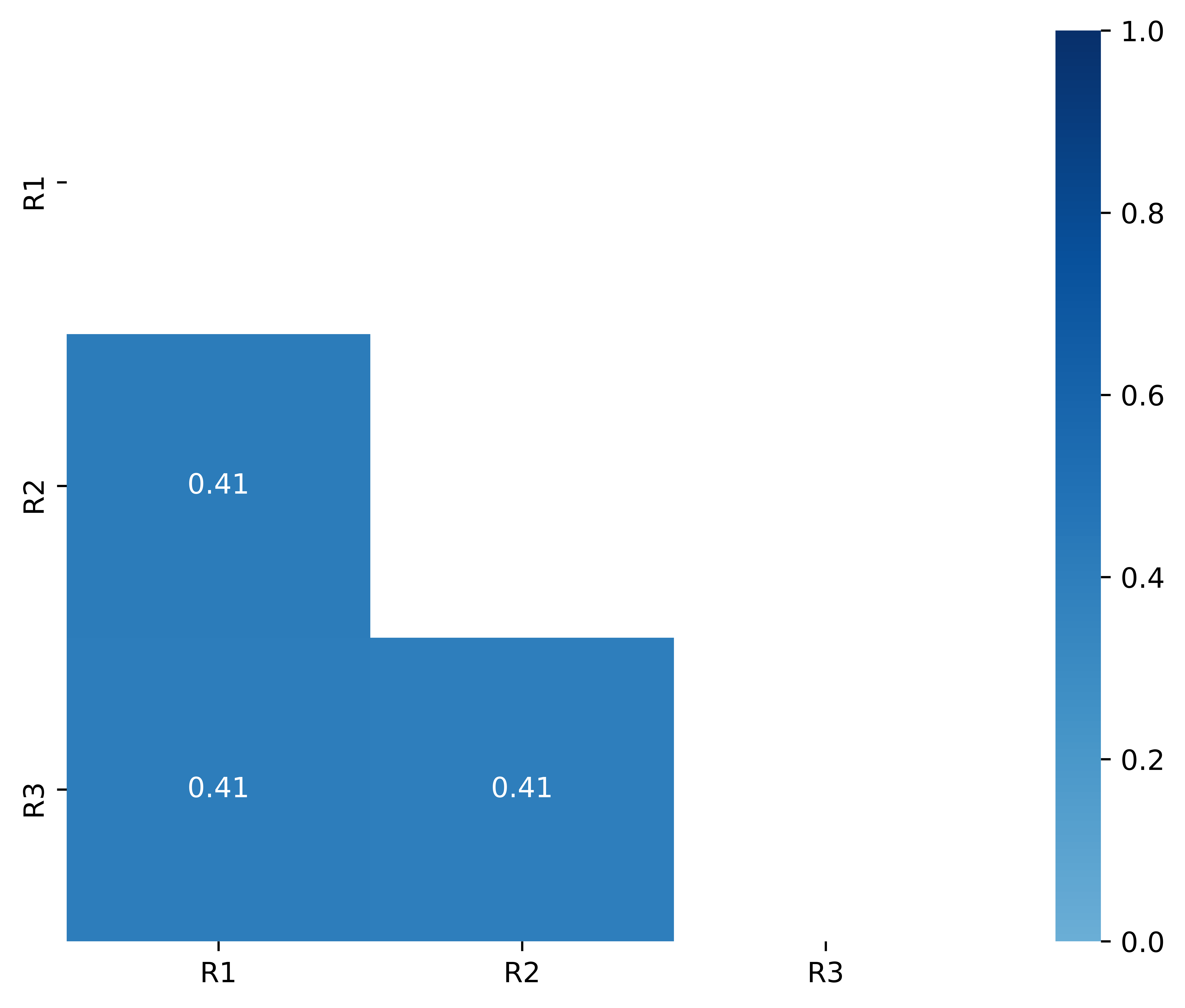}
    \caption{Cross-rater reliability analysis on the VCR\_P annotation task and datasets. Each cell in the heatmap represents the xRR score between pairwise sets of repetitions.}
    \label{fig:vcr_p_xrr}
\end{figure}

\begin{figure}[ht!]
    \centering
    \includegraphics[width=0.32\textwidth]{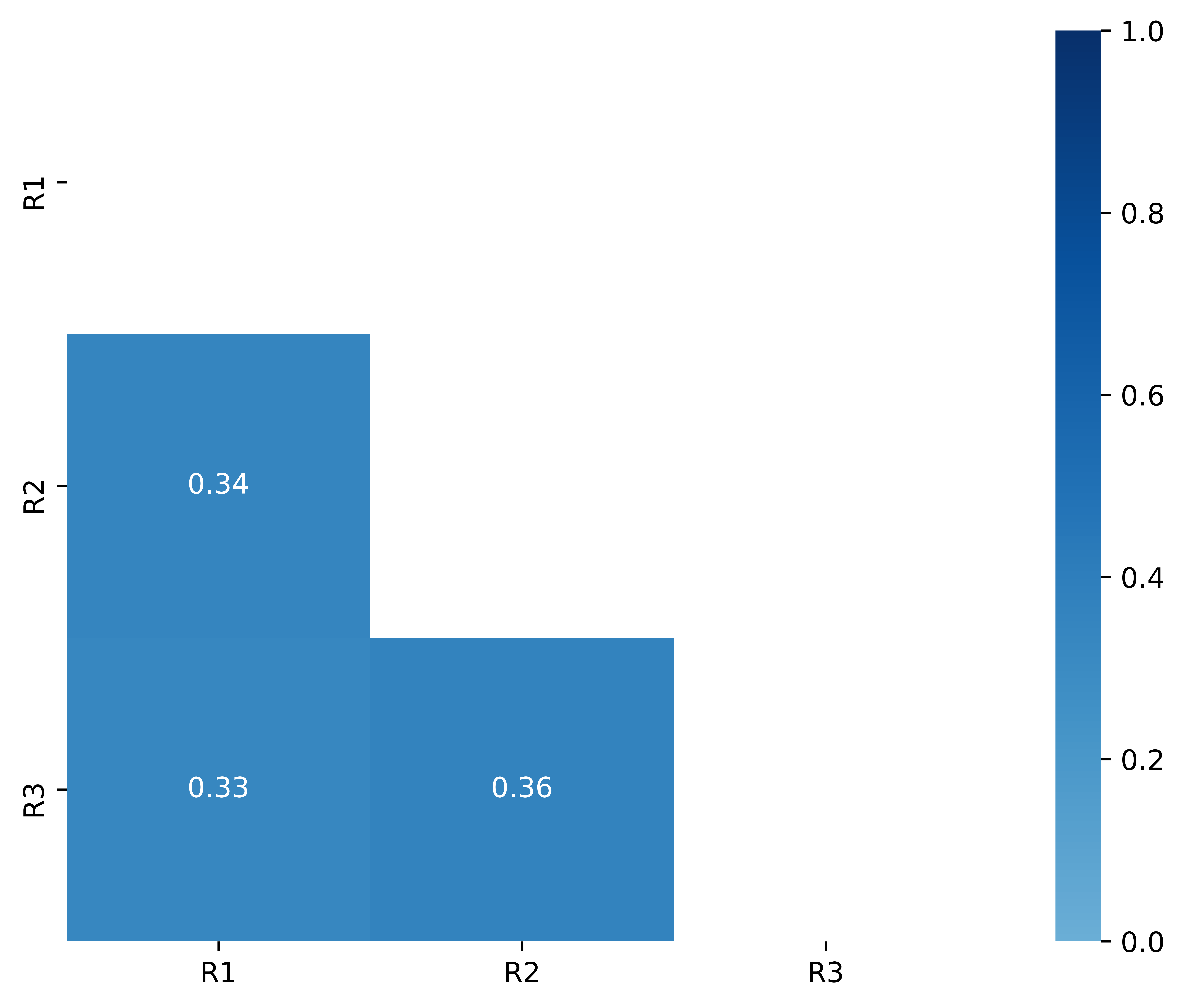}
    \caption{Cross-rater reliability analysis on the VCR\_L annotation task and datasets. Each cell in the heatmap represents the xRR score between pairwise sets of repetitions.}
    \label{fig:vcr_l_xrr}
\end{figure}

\begin{figure}[ht!]
    \centering
    \includegraphics[width=0.32\textwidth]{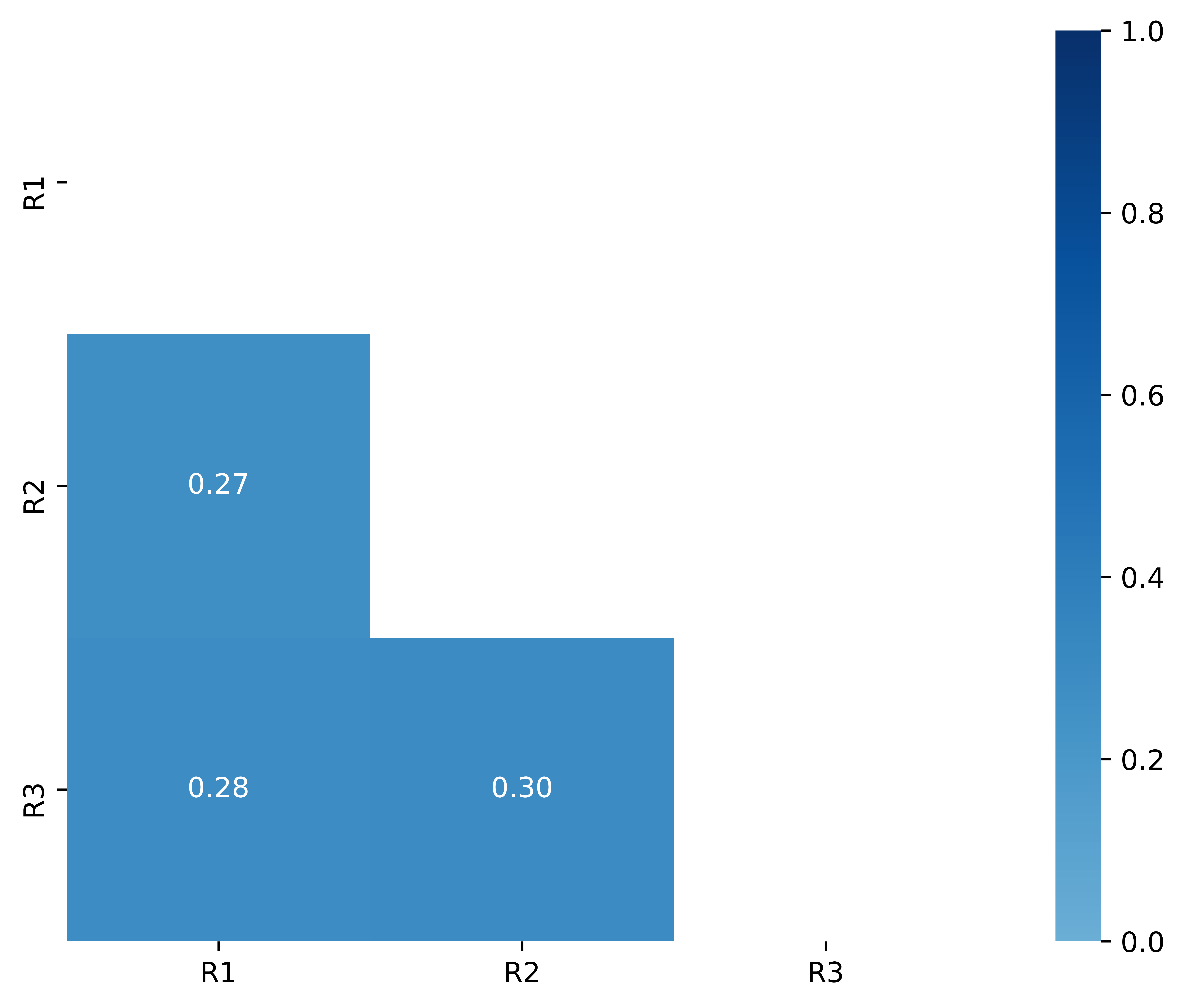}
    \caption{Cross-rater reliability analysis on the VCR\_O annotation task and datasets. Each cell in the heatmap represents the xRR score between pairwise sets of repetitions.}
    \label{fig:vcr_o_xrr}
\end{figure}

\begin{figure}[ht!]
    \centering
    \includegraphics[width=0.32\textwidth]{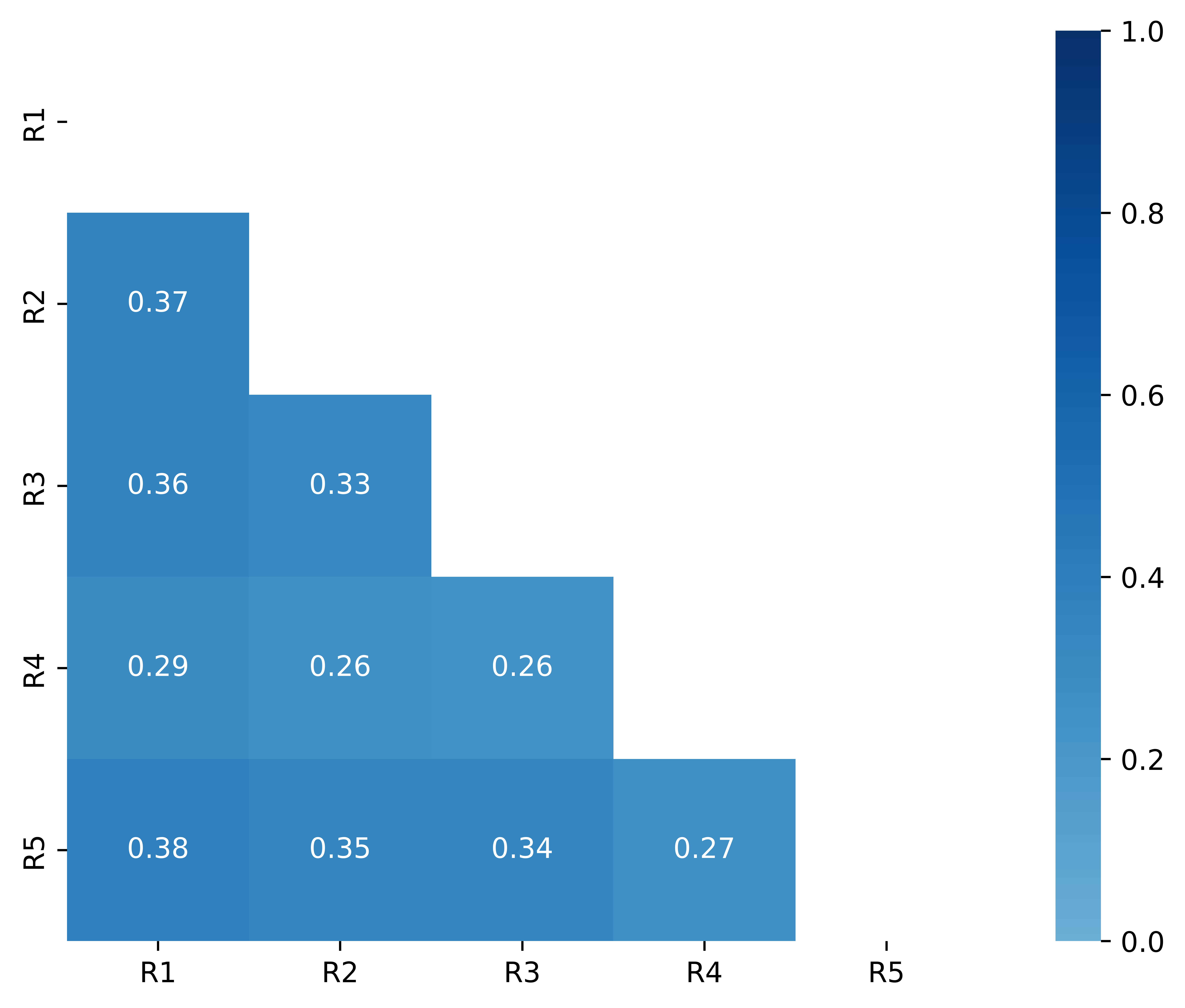}
    \caption{Cross-rater reliability analysis on the PR annotation task and datasets. Each cell in the heatmap represents the xRR score between pairwise sets of repetitions.}
    \label{fig:pr_xrr}
\end{figure}

\begin{figure}[ht!]
    \centering
    \includegraphics[width=0.32\textwidth]{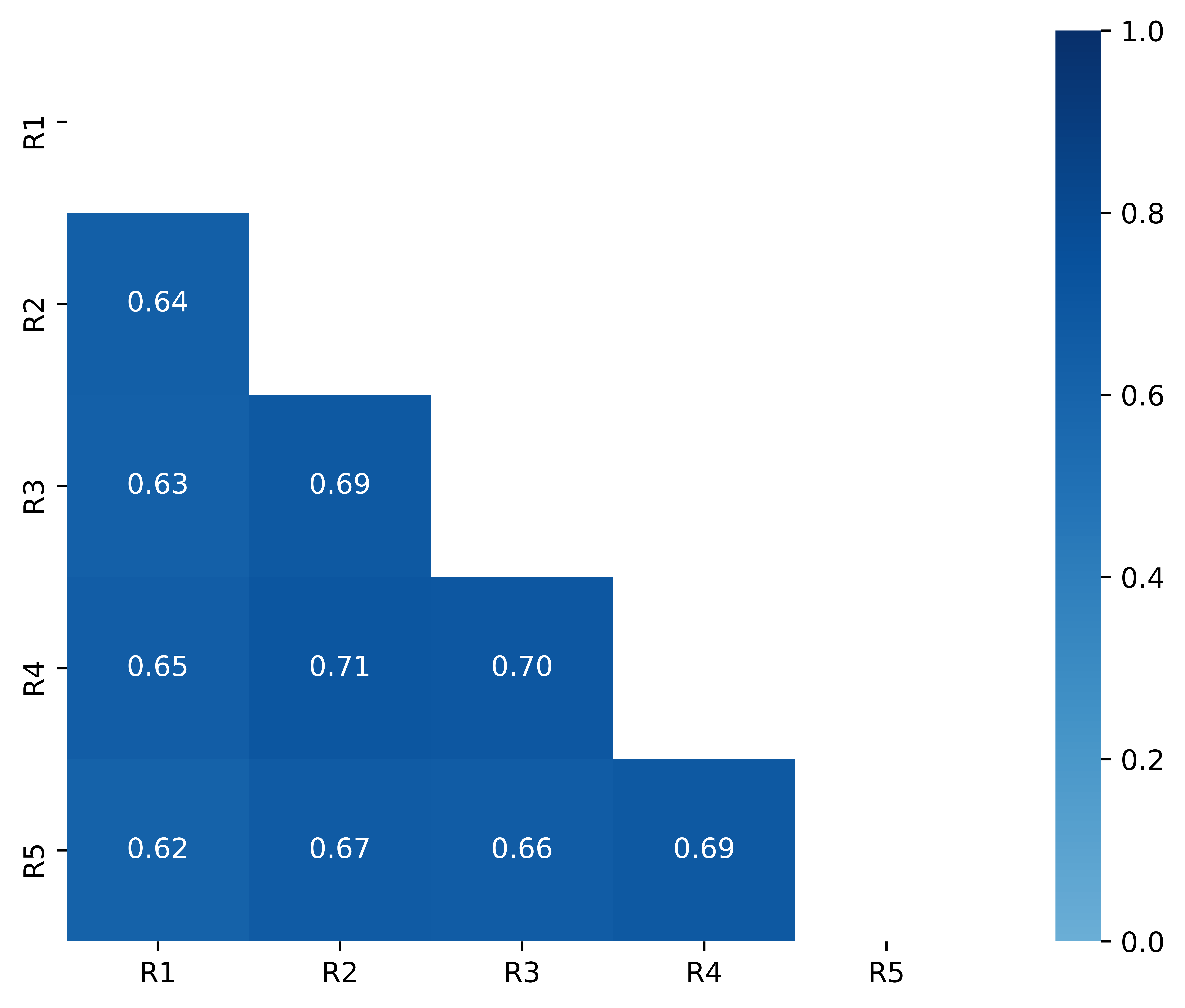}
    \caption{Cross-rater reliability analysis on the CT annotation task and datasets. Each cell in the heatmap represents the xRR score between pairwise sets of repetitions.}
    \label{fig:ct_xrr}
\end{figure}

\begin{figure}[ht!]
    \centering
    \includegraphics[width=0.38\textwidth]{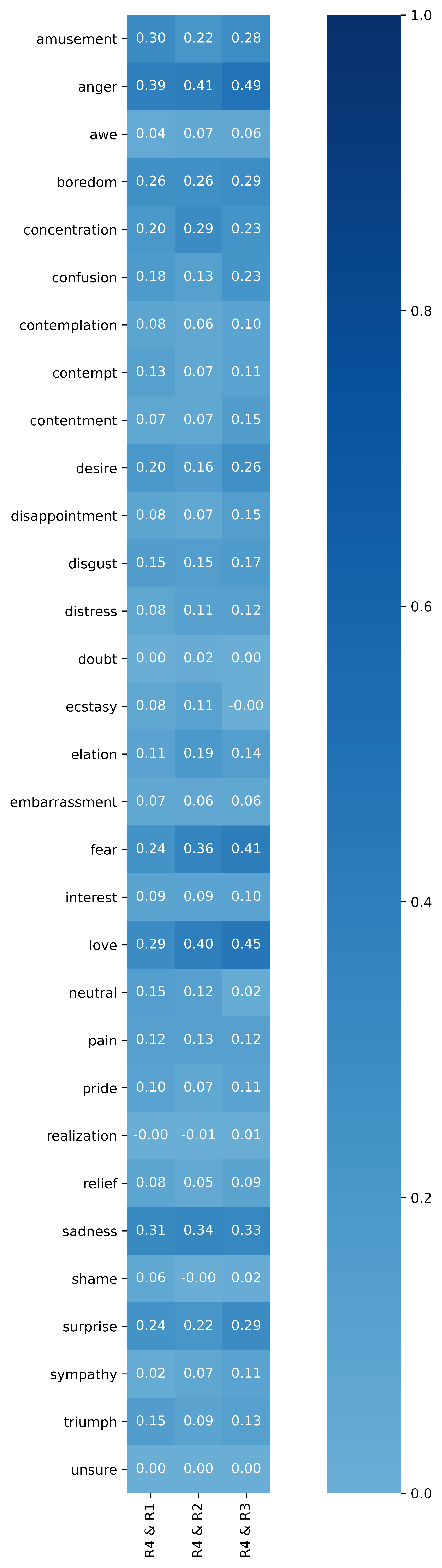}
    \caption{Cross-rater reliability analysis on $R4$ of the IRep annotation task and dataset. Each cell in the heatmap represents the xRR score between pairwise sets of repetitions.}
    \label{fig:irep_xrr}
\end{figure}

In this section, we report on the replicability similarity analysis, which indicated the degree of agreement between two rater pools, thus between two repetitions of the same task. This analysis is performed using the cross-replication reliability metric on the video concept relevance tasks (VCR\_ALL in Figure~\ref{fig:vcr_all_xrr}, VCR\_E in Figure~\ref{fig:vcr_e_xrr}, VCR\_P in Figure~\ref{fig:vcr_p_xrr}, VCR\_L in Figure~\ref{fig:vcr_l_xrr}, VCR\_O in Figure~\ref{fig:vcr_o_xrr}), product reviews task (in Figure~\ref{fig:pr_xrr}), crisis tweets task (in Figure~\ref{fig:ct_xrr}), and video human facial expressions task (in Figure~\ref{fig:irep_xrr}).

\end{document}